\documentclass[lettersize,journal]{IEEEtran}
\usepackage{amsmath,amsfonts}
\usepackage{algorithmic}
\usepackage{algorithm}
\usepackage{array}
\usepackage[caption=false,font=normalsize,labelfont=sf,textfont=sf]{subfig}
\usepackage{textcomp}
\usepackage{stfloats}
\usepackage{url}
\usepackage{verbatim}
\usepackage{graphicx}
\usepackage{cite}
\usepackage{times}
\usepackage{epsfig}
\usepackage{graphicx}
\usepackage{amsmath}
\usepackage{amssymb}
\usepackage{booktabs}
\usepackage{enumitem}
\usepackage{soul}
\usepackage[normalem]{ulem}
\usepackage{color}
\usepackage{cite}
\usepackage{multirow}
\usepackage{makecell}
\usepackage{xr}
\usepackage{color}
\usepackage{hyperref} 
\begin{document}

\title{RefComp: A Reference-guided Unified Framework for Unpaired Point Cloud Completion}

\author{
Yixuan Yang,
Jinyu Yang,~\IEEEmembership{Graduate Student Member,~IEEE}, 
Zixiang Zhao,
Victor Sanchez$^{*}$,~\IEEEmembership{Member,~IEEE},\\
\thanks{$^*$Corresponding author.}

Feng Zheng$^{*}$,~\IEEEmembership{Member,~IEEE}

\thanks{Yixuan Yang is with the Southern University of Science and Technology, Shenzhen, China, and also with the University of Warwick, Coventry, U.K. (e-mail: arnoldyang97@gmail.com).}
\thanks{Jinyu Yang is with Tapall.ai. (e-mail: jinyu.yang96@outlook.com).}
\thanks{
Zixiang Zhao is with the Photogrammetry and Remote Sensing, ETH Z\"urich,  8093 Z\"urich, Switzerland (E-mail: zixiang.zhao@ethz.ch)
}
\thanks{Victor Sanchez is with the University of Warwick, Coventry, U.K. (e-mail: v.f.sanchez-silva@warwick.ac.uk).}
\thanks{Feng Zheng is with Southern University of Science and Technology, Shenzhen, China (e-mail: f.zheng@ieee.org).}



\markboth{Journal of \LaTeX\ Class Files,~Vol.~14, No.~8, August~2021}
{Shell \MakeLowercase{\textit{et al.}}: A Sample Article Using IEEEtran.cls for IEEE Journals}
}

\maketitle

\begin{abstract}
The unpaired point cloud completion task aims to complete a partial point cloud by using models trained with no ground truth.
Existing unpaired point cloud completion methods are class-aware, i.e., a separate model is needed for each object class.
Since they have limited generalization capabilities, these methods perform poorly in real-world scenarios when confronted with a wide range of point clouds of generic 3D objects.
In this paper, we propose a novel unpaired point cloud completion framework, namely the Reference-guided Completion (\textit{RefComp}) framework, which attains strong performance in both the class-aware and class-agnostic training settings.  
The \textit{RefComp} framework transforms the unpaired completion problem into a shape translation problem, which is solved in the latent feature space of the \textit{partial} point clouds. To this end, 
we introduce the use of \textit{partial-complete} point cloud pairs, which are retrieved by using the partial point cloud to be completed as a template. 
These point cloud pairs are used as reference data to guide the completion process.
Our \textit{RefComp} framework uses a reference branch and a target branch with shared parameters for shape fusion and shape translation via a Latent Shape Fusion Module (LSFM) to enhance the structural features along the completion pipeline. 
Extensive experiments demonstrate that the \textit{RefComp} framework achieves not only state-of-the-art performance in the class-aware training setting but also competitive results in the class-agnostic training setting on both virtual scans and real-world datasets. 
\end{abstract}


\section{Introduction}

Point clouds are increasingly used in a wide range of applications, including autonomous driving \cite{3ddetection1,3ddetection2,3ddetection3,3ddetection4,mm_wang2024virpnet,mm_an2024esc}, 3D object reconstruction \cite{3dre2,3dre3, ponder,mm_xu2024_reconstruction,li2023hierarchical_recon}, and heritage restoration \cite{croce2021semantic,grilli2019classification}. 
Due to the limitations in sensor performance and occlusions, the collected point cloud data are often incomplete and noisy.
Such incomplete and noisy data may hinder the performance of many downstream tasks that rely on complete point clouds, \textit{e.g.}, 3D classification and segmentation \cite{pointnet,mm_zhang2024pointgt}.
Therefore, point cloud completion is an essential task for several of these downstream tasks. 

Recently, many point cloud completion methods \cite{sup_dai2017shape,sup_huang2020pf,sup_liu2020morphing, sup_tchapmi2019topnet,sup_wang2020cascaded,sup_wen2020point,sup_xie2020grnet,sup_zhang2020detail,yu2021pointr,MM_completion} have been proposed within the \textit{paired} partial-complete point cloud setting thanks to advances in processing point clouds directly\cite{PointBert,qi2017pointnet++,yang2018foldingnet,tcsvt2022psnet}. 
This setting relies on pairs of partial and corresponding complete point clouds during training. 
Despite their strong performance, methods proposed in this setting may not generalize well as it is hard to collect large-scale datasets 
for the wide range of objects that may be found in real-world scenarios.
Thus, designing completion methods that rely on \textit{unpaired} 
partial point cloud data is gaining attention \cite{pc_unpaired_pcl2pcl}. 
In other words, methods that require no ground truth for the partial point clouds used during training.
Building upon this foundation, existing methods predominantly employ the class-aware completion setting.
Specifically, these approaches necessitate distinct training models for different object classes, significantly increasing the costs associated with training, storage, and deployment. 
More critically, such models face substantial limitations in real-world applications, as they require users to first identify the object class before selecting the appropriate model for completion. 
To address these challenges, we propose a novel class-agnostic point cloud completion framework, to enhance the applicability and utility of point cloud completion in real-world scenarios.

\begin{figure}[!t]
	\centering
	\includegraphics[width=0.98\linewidth]{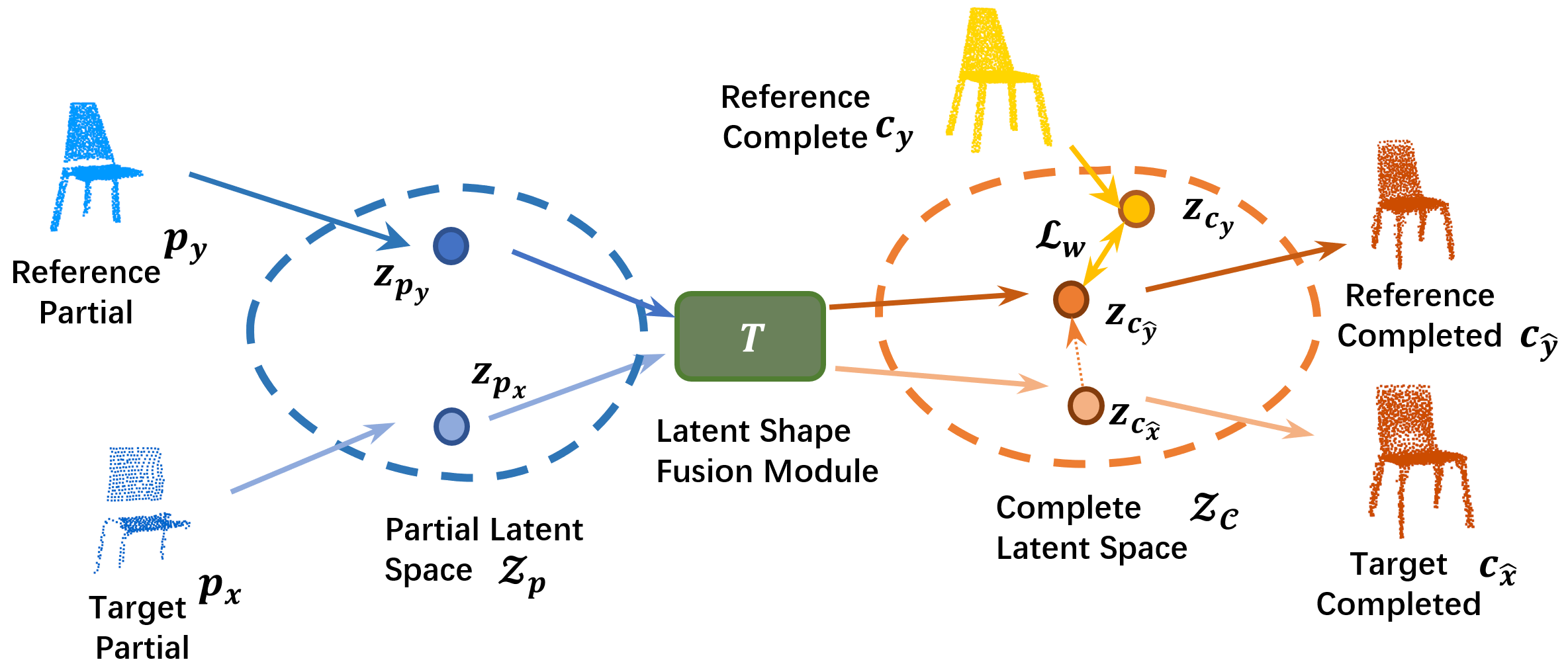}
	\caption{\textbf{Unpaired point cloud completion by shape translation.} $c_y$ is corrupted  to create $p_y$ by using $p_x$ as the corruption template. $z_{p_y}$ and $z_{p_x}$ are mapped into features representing complete point clouds by the Latent Shape Fusion Module (LSFM), $T$. $z_{c_{\hat{y}}}$ is forced to be in the $Z_\mathcal{C}$ space by a Wasserstein Distance loss, which \textit{pulls} $z_{c_{\hat{x}}}$ into the complete latent space. 
 Finally, the completed features are decoded back to the complete point clouds $c_{\hat{x}}$ and $c_{\hat{y}}$.}
	\label{shape_translation}
  \vspace{-1.5em}
\end{figure}

Current unpaired point cloud completion methods predominantly utilize complete point cloud models or samples as prior knowledge, employing generative models (such as GANs~\cite{pc_cycle4completion,pc_unpaired_pcl2pcl,pc_latent,pc_mpc} or Diffusion models~\cite{zhang2024reverse2complete}) to perform the completion task.
For instance, the state-of-the-art (SOTA) model Latent. \cite{pc_latent}, augments the training of a GAN by masking partial point clouds, enabling the model to learn more structural features.
The Inversion. \cite{pc_shape_inversion} model degrades complete point clouds to create partial point clouds and employs a pre-trained GAN for conditional completion.
The Cycle. \cite{pc_cycle4completion} model, which is inspired by CycleGAN \cite{CycleGAN}, uses a GAN to impose cycle supervision on the completion process.
The Energy~\cite{cui2022energy}, similarly employs the discriminator to quantify the domain discrepancy between completed point clouds and complete point clouds.
In contrast to the aforementioned methods that employ GANs for discrimination.
R2C~\cite{zhang2024reverse2complete} utilizes a diffusion model for point cloud completion. This method, due to the inherent characteristics of the diffusion model, is computationally complex and suffers from slow inference speed.
Moreover, its experimental performance does not show any advantage compared to other methods, leaving substantial room for improvement.
On the other hand, P2C~\cite{cui2023p2c} introduces a self-supervised point cloud completion approach that can effectively complete point clouds without relying on complete point clouds as prior information.


While the aforementioned methods have demonstrated promising results, they are primarily confined to class-aware settings and have not fully explored the potential of class-agnostic frameworks for unpaired completion tasks.
In this work, we propose a novel reference-guided framework that supports both class-aware and class-agnostic unpaired point cloud completion, significantly enhancing its generalization capability for real-world applications.
Our unified framework, termed \textit{\underline{Ref}erence-guided \underline{Comp}letion (RefComp)}, reformulates the completion task as a shape translation problem~\cite{cui2022energy}. The framework retrieves a complete reference point cloud from a multi-class dataset with the target partial input as a template, while generating the corresponding partial reference through degradation.
As illustrated in Fig. \ref{shape_translation}, RefComp employs an encoder-decoder architecture with a novel Latent Shape Fusion Module (LSFM) to perform shape translation. 
The LSFM effectively integrates missing structural features into the target partial point cloud's features, which are then aligned with the reference complete point cloud's feature space. Finally, these features are decoded back into the point cloud space.

We evaluate our \textit{RefComp} framework on widely recognized point cloud completion benchmarks using both class-aware and class-agnostic training settings. These benchmarks include virtual scan datasets (EPN-3D \cite{dai20173D-EPN}, CRN \cite{wang2020CRN}, PartNet \cite{mo2019partnet}) and real-world datasets (KITTI \cite{Geiger2013KITTI}, ScanNet \cite{dai2017scannet}, and MatterPort3D \cite{chang2017matterport3d}). In comparison to class-aware models \cite{pc_latent,pc_shape_inversion,pc_cycle4completion,pc_unpaired_pcl2pcl}, our framework exhibits competitive performance as a class-agnostic model. Furthermore, when trained in a class-aware setting, our framework achieves SOTA performance across the classes. Our main contributions are summarized as follows:

\begin{itemize}
\setlength{\itemsep}{1pt}
\setlength{\parsep}{0pt}
\setlength{\parskip}{0pt}
\item To the best of our knowledge, \textit{RefComp} is the first unified framework for class-agnostic unpaired point cloud completion, which makes the completion pipeline more practical in the real-world scenario.

\item The \textit{RefComp} framework transforms the completion problem into a shape translation problem with the retrieved reference data.

\item The \textit{RefComp} framework introduces a novel module, called LSFM, to fuse the missing structural features into the features of the target partial point cloud to improve the completion results.

\item The \textit{RefComp} framework achieves SOTA performance in the class-aware training setting and competitive performance in the class-agnostic training setting on both virtual scan and real-world datasets.
\end{itemize}

\section{Related Work}
We first discuss current paired and unpaired partial point cloud completion methods.
Next, we focus on the reference-guided models applied in other computer vision tasks and discuss the rationale behind introducing this idea to the unpaired point cloud completion task.

\subsection{Point Cloud Completion} \label{related_work:1}
Before PointNet\cite{pointnet}, the majority of point cloud completion methods used to rely on 3D voxel grids \cite{voxel_dai2017shape,voxel_wang2017shape,voxeL_yang20173d}. PointNet is the first deep-learning method to process unordered point clouds directly.
The Point Completion Network (PCN)~\cite{yuan2018pcn} proposed after PointNet, uses PointNet as the backbone network to first complete the point cloud and subsequently FoldingNet~\cite{yang2018foldingnet} to refine it. 
After these seminal methods, several paired partial point cloud completion methods have been proposed, i.e., methods that are fully supervised by the ground truth \cite{sup_dai2017shape,sup_huang2020pf,sup_liu2020morphing, sup_tchapmi2019topnet,sup_wang2020cascaded,sup_wen2020point,sup_xie2020grnet,sup_zhang2020detail,yu2021pointr,MM_completion}. 

Although fully supervised methods have demonstrated impressive results, they have limitations to generalize to real-world scenarios. This is because it can be challenging to collect the required ground truth data for the wide range of objects that may appear in a specific context. 
Consequently, a new wave of methods utilizing unpaired point clouds for training has emerged. These approaches aim to leverage complete point clouds that are unpaired with the partial inputs as prior knowledge for completion tasks, as exemplified by \cite{pc_latent,pc_unpaired_pcl2pcl, pc_shape_inversion, pc_cycle4completion, cui2022energy, zhang2024reverse2complete}. 
Such methods continue to gain significant attention and remain highly prevalent in current research.

Pcl2pcl\cite{pc_unpaired_pcl2pcl} is the first method to tackle the unpaired partial point cloud completion task.
It uses a pre-trained partial point cloud encoder and a complete point cloud encoder to extract latent features. It then uses a GAN to constrain the shape of the completed point cloud by distinguishing between actual complete point clouds and those completed from partial point clouds. 
Cycle. \cite{pc_cycle4completion}, which is based on \cite{pc_unpaired_pcl2pcl} and inspired by CycleGAN \cite{CycleGAN}, uses a bidirectional mapping between complete and partial point clouds to capture the relationship between the shapes.
Inversion. \cite{pc_shape_inversion} applies a GAN Inversion method, which uses a fixed pre-trained generator to map the partial point cloud to a complete one by acquiring \emph{a priori} shape information.
The Latent. \cite{pc_latent}, uses a downsampling module to obtain several partial point clouds at different resolutions and then maps them into a unified latent feature space for completion by also learning from occlusions.
R2C~\cite{zhang2024reverse2complete} adopts the diffusion model for point cloud completion, utilizing both the partial input point cloud and a reference complete point cloud as prior knowledge during the reverse process. However, this method has only been evaluated on three object categories, and the diffusion model still heavily relies on extensive data priors. Moreover, it requires significantly more computational resources and time for both training and inference compared to conventional models. These limitations indicate that point cloud completion using this approach remains underdeveloped and warrants further work.

Recent advancements have introduced self-supervised approaches for point cloud completion, notably exemplified by P2C~\cite{cui2023p2c}. However, similar to the aforementioned unpaired methods, these approaches primarily focus on class-aware training settings, which significantly limits their applicability in real-world tasks. To address this limitation, we aim to develop a novel framework capable of performing both class-aware and class-agnostic point cloud completion.
Our proposed solution introduces a retrieval-based mechanism that leverages partial point clouds to identify geometrically similar complete point clouds within the dataset. These retrieved complete point clouds serve a dual purpose: (1) as reference conditions for shape completion, and (2) as implicit pseudo-labels that guide the partial point cloud towards completion with plausible category-specific shapes.

\begin{figure*}[!t]
	\centering
	\includegraphics[width=1\linewidth]{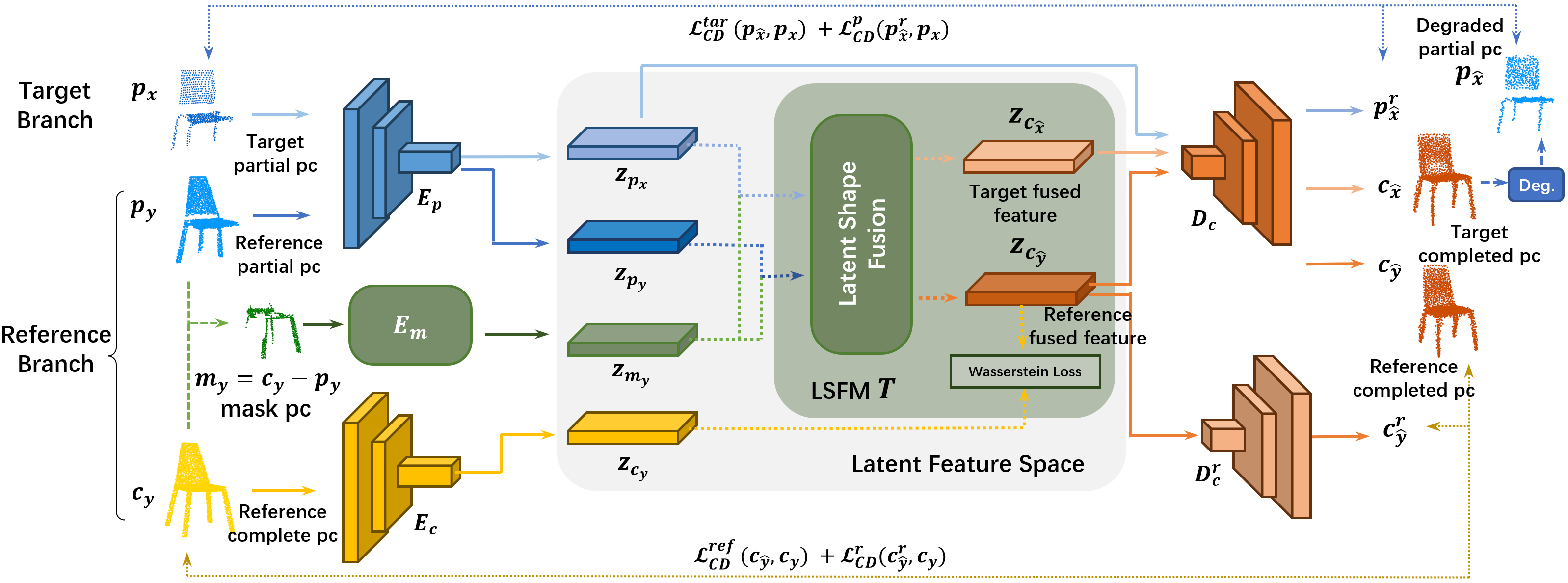}
	\caption{\textbf{Our \textit{RefComp} framework.}
The reference branch shares the parameters with the target branch to assist in the unpaired point cloud completion. $p_x, p_y$ are encoded by the same encoder $E_p$.  Completion is achieved by the Latent Shape Fusion Module (LSFM) with the mask point cloud} $m_y$ in the latent feature space. $c_y$ provides the alignment objective, ensuring the completed features $z_{c_{\hat{y}}}$ align with the complete features $z_{c_y}$. A 
 decoder with shared parameters decodes  $z_{c_{\hat{x}}}$ and $z_{c_{\hat{y}}}$ to complete point clouds. The shape constraints are imposed on  $p_{\hat{x}}$ and $p_x$, as well as on $c_{\hat{y}}$ and $c_y$.
\label{mapping_jpg}
  \vspace{-0.8em}
\end{figure*}

\subsection{Reference Models in Vision Tasks} 
Reference-guided models have gained popularity in many tasks, including image super-resolution \cite{reference_sr,reference_masa,reference_crossnet}, image style-transfer \cite{ref_bringing,ref_hairclip}, and image de-raining \cite{ref_deraining}. They are also used to enhance cross-modality models. 
For example, the works in  \cite{ReferringSeg, gres} employ text as the reference to describe the objects that need to be segmented accurately.
More recently,  the \emph{visual prompt} concept has emerged as another example of the advantages of reference models \cite{visual_prompt, imagesinimages,fang2023pc_prompt}. In essence, this concept involves using paired images and point clouds as prompts to facilitate 
several tasks. For example,  in \cite{fang2023pc_prompt}, researchers propose in-context prompt learning to solve multiple tasks related to point clouds, i.e., upsampling, registration, part segmentation, and denoising. That particular work is based on the Transformer model, which relies on positional encoding. 
When partial point clouds are extremely sparse, i.e.,  severely incomplete, the positional encoding of the partial parts cannot provide sufficient information to accurately complete the point cloud. Hence, in such cases, this Transformer-based method fails to produce strong results for the unpaired partial point cloud completion task.

In general, reference models are rarely used in unpaired point cloud completion tasks. To the best of our knowledge,  only MPC\cite{pc_mpc} uses reference data in a diversity completion context, but it still relies on the GAN's capabilities. Although other works\cite{pc_latent,pc_cycle4completion,pc_shape_inversion} use complete point clouds from specific datasets (i.e., ShapeNet), they do not use these data explicitly as the reference in the way our \textit{RefComp} framework does.
These methods only discriminate between the distribution of actual complete point clouds and that of the completed ones by using a GAN. 

\section{Proposed Framework}\label{Method}
Our \textit{RefComp} framework uses reference data to impose shape constraints on the target partial point cloud. 
We first discuss how to generate and retrieve such reference data
(Section \ref{con_ref_data}). 
Then, we explain how our framework transforms the unpaired point cloud completion problem into a shape translation problem in the latent feature space (Section \ref{shape_translation_section}). To this end, we introduce the LSFM to fuse missing structural features into the features of the target partial point cloud (Section \ref{LSFM}). 
Finally, in Sections \ref{decode_latent} and \ref{Shape constrain}, we discuss how to decode the fused features into a complete point cloud while constraining the shape generation. Our complete \textit{RefComp} framework is depicted in Fig. \ref{mapping_jpg}.

\subsection{Reference Data} \label{con_ref_data} 

\begin{figure}[!t]
	\centering
	\includegraphics[width=1\linewidth]{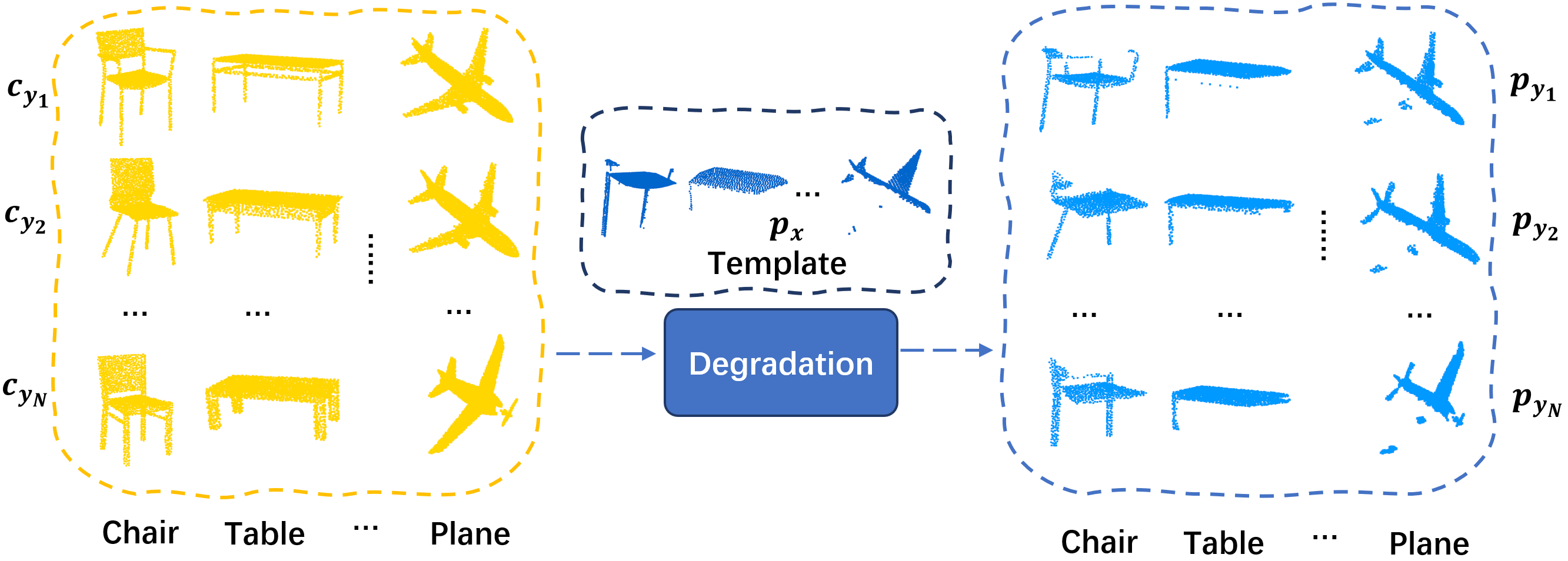}
	\caption{\textbf{Generation of the reference data.} %
{For each point in  $p_x$, the degradation module uses K-nearest neighbour (KNN)} to find the top K closest points in $c_y$, where  $p_y$ is the union of these points. A CD loss between each $p_y$ and  $p_x$ is used to find the top-N reference pairs.}
	\label{degard}
\end{figure}

Like other reference-guide tasks \cite{reference_sr,reference_masa,ref_bringing},  we use a dataset of complete point clouds to create reference pairs, denoted by $\{p_y, c_y\}$, where, $c_y$, is a complete point cloud and  $p_y$ is a partial point cloud generated by corrupting $c_y$. 
Specifically, we use the non-parameter degradation module proposed in \cite{pc_shape_inversion} to corrupt   $c_y$ by using $p_x$ as the corruption template, where $p_x$ is the target partial point cloud to be completed. 
Note that $p_x$ and $p_y$ depict different objects and that a single $p_x$ can be used as the corruption template to create several reference pairs, $\{p_y, c_y\}$. 
After creating the reference pairs, we use the Chamfer Distance (CD) between $p_x$ and all $\{p_y\}$  of the same class to select the top $N$ pairs, denoted by $\{ \{p_{y_1},c_{y_1}\},\dots, \{p_{y_N},c_{y_N}\} \}$, where $\{p_{y_n},c_{y_n}\}$ represents the $n^{th}$ reference pair with $n = 1\dots N$. 
This is shown in Fig. \ref{degard}. 

\subsection{{Shape Translation}}\label{shape_translation_section}

The shape translation in our \textit{RefComp} framework is between the partial point cloud space $\mathcal{P}$ and the complete point cloud space $\mathcal{C}$, as done in previous works \cite{pc_latent,pc_unpaired_pcl2pcl}.
However, compared to previous works, our shape translation involves multiple point clouds: the reference pairs $\{\{p_y, c_y\}\}$ and the target partial point cloud $p_x$, where $\big(p_x, \{p_y\}\big) \in \mathcal{P}$ and $\{c_y\} \in \mathcal{C}$. 
To map the point clouds into a latent feature space, we use two encoders $E_p$ and $ E_c$ on $\big(p_x, \{p_y\}\big)$ and $\{c_y\}$ to obtain $\big(z_{p_x}, \{z_{p_y}\}\big)$ and $\{z_{c_y}\}$, respectively.  Such mappings can then be formulated as: $E_p:\mathcal{P} \rightarrow Z_\mathcal{P}$ and $E_c:\mathcal{C} \rightarrow Z_\mathcal{C}$. Encoder 
$E_p$ computes the features for $p_x$ and  $\{p_y\}$ with a single set of shared parameters while ensuring that the features are \textit{aligned} in the latent feature space $Z_\mathcal{P}$. 
To translate the partial point cloud features to the complete point cloud features, we propose a completion module $T: Z_\mathcal{P} \rightarrow Z_\mathcal{C}$.
Finally, we use a shared-parameter decoder, denoted by $D_c: Z_\mathcal{C} \rightarrow \mathcal{C}$, to map the complete point cloud features back to the point cloud space, generating the complete point clouds $\big(c_{\hat{x}}, \{c_{\hat{y}}\}\big)$, where  $c_{\hat{x}}$ is the completed target point cloud. 

It is worth mentioning that $\{c_{\hat{y}}\}$ are completed based on the reference data, which allows computing the reconstruction loss with respect to the complete reference data $\{c_{{y}}\}$. Since the network parameters are fully shared, 
we can represent the mapping process from $p_x$ to $c_{\hat{x}}$ as follows:

\begin{equation}\label{target_mapping}
\begin{aligned}
c_{\hat{x}} =D_c\circ  T \circ E_p(p_x).
\end{aligned}
\end{equation}

Both encoders, $E_p$ and $ E_c$, adopt a two-layer PointNet\cite{pointnet}.  $E_p$ aims to capture the global representations of the partial point clouds. On the other hand, $E_c$ encodes the reference complete point cloud, which helps to align the partial point cloud features after completing them in the latent space. 
The decoder $D_c$ is a five-layer MLP.
The process of completing the reference partial point clouds $\{p_y\}$ is performed along a \emph{reference branch}. Note that this branch is supervised by the reference complete point clouds, $\{c_y\}$.
On the other hand, the completion process for our target partial point cloud, $p_x$, is performed along the \emph{target branch} with no supervision and by only leveraging prior knowledge from the reference branch.
To backpropagate gradients through the target branch and maintain the overall shape of  $p_x$, we use the $Deg(\cdot)$ module in \cite{pc_shape_inversion} to degrade $c_{\hat{x}}$ back to  $p_{\hat{x}}$. We then compute the CD-based loss between  $p_x$ and $p_{\hat{x}}$: 
\begin{equation}\label{Chamfer_distance}
\begin{aligned}
\mathcal{L}_{CD} (p_{\hat{x}}, p_x) &=  \frac{1}{\vert p_{\hat{x}}\vert}\sum_{a\in p_{\hat{x}}}\min_{b\in p_x}\Vert a-b\Vert_2^2 \\ &+
\frac{1}{\vert p_x\vert}\sum_{b\in p_x}\min_{a\in p_{\hat{x}}}\Vert a-b\Vert_2^2, 
\end{aligned}
\end{equation}
where $\{a, b\}$ are  points in  $p_{\hat{x}}$ and $p_x$, respectively. The same  CD-based loss is used along the reference branch. These losses can then be denoted as $\mathcal{L}_{CD}^{ref} (c_{\hat{y}}, c_y)$ and $\mathcal{L}_{CD}^{tar} (Deg(c_{\hat{x}}), p_x)$, for the reference and target branch, respectively.

As explained in Section \ref{con_ref_data}, we select $N$ reference pairs for each target partial point cloud, $p_x$. The \textit{RefComp} framework uses the $top-3$  pairs, $\{ \{p_{y_1},c_{y_1}\},\{p_{y_2},c_{y_2}\}, \{p_{y_3},c_{y_3}\} \}$,  as the reference data. In each training iteration, we randomly select a pair $\{p_y, c_y\} \in \{ \{p_{y_1},c_{y_1}\},\{p_{y_2},c_{y_2}\}, \{p_{y_3},c_{y_3}\} \}$,  which helps to enhance the diversity of the features used for completion.

\subsection{Latent Shape Fusion Module}\label{LSFM}
\begin{figure}[!t]
	\centering
	\includegraphics[width=1.02\linewidth]{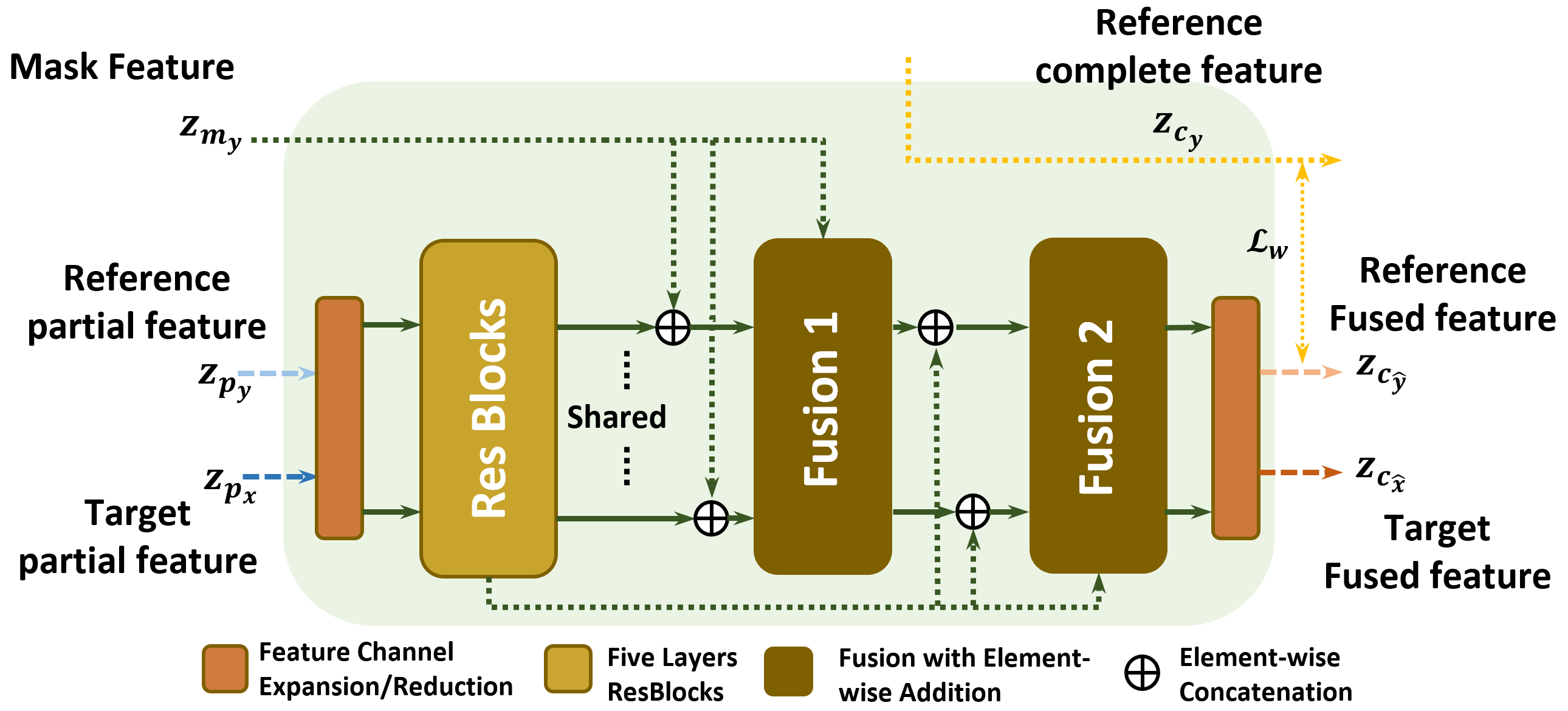}
	\caption{\textbf{The Latent Shape Fusion Module (LSFM).} 
The parameters between the reference and the target branch are shared to compute $z_{\hat{c}_x}$ and $z_{\hat{c}_y}$. To complete the partial point clouds in the latent feature space, the LSFM fuses the features of the partial point clouds, i.e., $z_{p_x}$ and $z_{p_y}$, with  $z_{m_y}$, which represents missing structural features}. 
	\label{Tf}
  \vspace{-0.8em}
\end{figure}

The shape translation problem in \textit{RefComp} is implemented by the completion module, $T$ (see Fig. \ref{mapping_jpg}). Specifically, this module is a \textit{Latent Shape Fusion Module} (LSFM) that fuses the features of the reference pairs into those of the target partial point cloud.
As depicted in Fig. \ref{Tf}, the LSFM comprises two branches: the reference branch to map $\{z_{p_y}\}$ to $\{z_{c_{\hat{y}}}\}$ and the target branch to map $z_{p_x}$ to $z_{c_{\hat{x}}}$, where $z_{c_{\hat{x}}}$ is the feature representation of the completed target point cloud $c_{\hat{x}}$.

As mentioned in Section \ref{con_ref_data}, each $p_y$ is generated by corrupting the corresponding $c_y$. Assuming that $p_y$ is occluded by a mask, the occluded region mask point cloudcan be calculated by $m_y = c_y - p_y$. 
The reference branch of the LSFM fuses $z_{p_y}$ with the mask $m_y$'s embedding, \textit{i.e.},  $z_{m_y}$, where $z_{m_y}$ is encoded by a separate encoder, $E_m$, that has the same architecture as $E_p$. 
The fused features are then mapped to $z_{c_{\hat{y}}}$, which represents the features of the reference complete point cloud, $c_y$. The fusion and mapping operation along the reference branch can then be represented as $z_{c_{\hat{y}}} = T(z_{p_y}, z_{m_y})$.

Since a similar corruption is expected to exist between $p_x$ and its associated $\{p_y\}$, the masks $\{m_y\}$ should also be useful to represent the missing parts in $p_x$. Based on this rationale, we exploit the parameter-sharing mechanism of the network and transfer the completion capability of the reference branch into the target branch. 
We employ feature fusion on $z_{m_y}$ and $z_{p_x}$, which can be represented as $z_{c_{\hat{x}}} = T(z_{p_x}, z_{m_y})$.
This means that the target branch and the reference branch use the same network architecture and share the model parameters. 

The architecture of the LSFM is shown in Fig. \ref{Tf}. It first expands the number of feature channels of the two inputs and subsequently uses a five-layer ResBlocks \cite{resnet} to capture additional low-level features. To incorporate structural features into the target branch, it concatenates the low-level features with $z_{m_y}$. It then applies two Fusion blocks to enhance the features extracted. 
n the Fusion1 block, we first process the fused features through a residual block and then perform an element-wise addition with the mask feature. After concatenating the resulting features with the partial features, in the Fusion2 block, we only perform an element-wise addition between the newly fused features and the previously computed partial features.
Finally, the number of feature channels is reduced to get the completed fused feature $z_{c_{\hat{x}}}$ and $z_{c_{\hat{y}}}$.
We formalize this latent shape fusion process as follows (channel dimension changes are omitted): 
\begin{equation}
\scalebox{0.88}{$
\begin{aligned}
    z_{c_{\hat{x}}} &= \left[\left[R(R^5(z_{p_x}) \oplus z_{m_y})+z_{m_y}\right] \oplus R^5(z_{p_x})\right]+ R^5(z_{p_x}), \\
    z_{c_{\hat{y}}} &=\left[\left[R(R^5(z_{p_y}) \oplus z_{m_y})+z_{m_y}\right] \oplus R^5(z_{p_y})\right]+ R^5(z_{p_y}),
\end{aligned}
$}
\end{equation}
where $R$ is the residual block, $R^5$ denotes five consecutive residual blocks, $\oplus$ denotes  concatenation, and $+$ denoted element-wise addition. Each residual block is implemented as a 1$\times$1 convolution layer.

In each training iteration, a randomly selected  $p_y$ from the set of reference pairs associated with $p_x$ is used. 
This strategy allows computing diverse $\{m_y$\}, which can enhance the learning capability of LSFM as it facilitates extracting a diverse range of structural features. This strategy also helps to improve the generalization capabilities of the target branch.

Note that the LSFM acts as a bridge between the partial point cloud latent space $Z_\mathcal{P}$ and the complete point cloud latent space $Z_\mathcal{C}$. To further guarantee that the fused features are \textit{bridged} appropriately into $Z_\mathcal{C}$, we use the Wasserstein distance \cite{wasserstein} to define the loss function $\mathcal{L}_{\mathcal{W}}$. This loss forces $z_{c_{\hat{y}}} \sim Z_{\hat{\mathcal{C}}} $ to be close to $z_{c_y} \sim Z_{\mathcal{C}}$ , where $z_{c_y}$ is computed  by $E_c$. 
Our loss $\mathcal{L}_{\mathcal{W}}$ is formulated as:
\begin{equation}\label{Wasserstein_distance}
\begin{aligned}
 \mathcal{L}_{\mathcal{W}}(Z_{\hat{\mathcal{C}}},Z_{\mathcal{C}}) = \inf_{\gamma\in\Pi(Z_{\hat{\mathcal{C}}},Z_{\mathcal{C}})} \mathbb{E}_{(z_{c_{\hat{y}}}, z_{c_y})\sim\gamma}\parallel z_{c_{\hat{y}}}-z_{c_y} \parallel, 
\end{aligned}
\end{equation}
where $\Pi$ is the joint distribution of $Z_{\hat{\mathcal{C}}}$ and $Z_{\mathcal{C}}$.
Owing to the LSFM's parameter-sharing mechanism, when the reference branch translates $z_{p_y}$ into the complete latent space $Z_{\mathcal{C}}$, the translation result of the target branch, \textit{i.e.}, $z_{c_{\hat{x}}}$, is \textit{pulled} into $Z_{\mathcal{C}}$.

\subsection{Decoding Latent Features}\label{decode_latent}

To decode $z_{c_{\hat{x}}}$ and $z_{c_{\hat{y}}}$ from the complete point cloud latent feature space to the point cloud space, we leverage a five-layer MLP, $D_c$, as the decoder  (see Fig. \ref{mapping_jpg}). 
The decoder $D_{{c}}$ also shares parameters between the reference and target branch. The loss function for these two branches are $\mathcal{L}_{CD}^{ref} (c_{\hat{y}}, c_y)$ and $\mathcal{L}_{CD}^{tar} (Deg(c_{\hat{x}}), p_x)$, respectively, as described in Section \ref{shape_translation_section}.

To further guarantee the plausibility of the completed shapes, we use an independent decoder $D_{c}^{r}$ to decode $z_{c_{\hat{y}}}$ into $c_{\hat{y}}^{r}$, which helps the LSFM to learn more structural features and subsequently facilitate the generation of more uniform surfaces in the point clouds. 
To strengthen $D_{c}$'s capability of learning structural properties from the partial point cloud, we map $z_{p_x}$ into $p_{\hat{x}}^r$ (see Fig. \ref{mapping_jpg}) directly by using $D_{c}$. This strategy allows $D_{c}$ to learn corruption patterns that can help to enhance the details of the completed target point cloud and thus, improve the overall performance of 
the target branch.
The shape generation is supervised by the losses $\mathcal{L}_{CD}^{r}(c_{\hat{y}}^{r}, c_y)$ and $\mathcal{L}_{CD}^{p}(p_{\hat{x}}^r, p_x)$. 
The loss functions for the reference and target branches are, respectively,  $\mathcal{L}_{CD}^{ref} + \mathcal{L}_{CD}^{r}$ and $\mathcal{L}_{CD}^{tar} + \mathcal{L}_{CD}^{p}$. The overall loss function of the \textit{RefComp} framework is then:
\begin{equation}\label{overall loss}
\begin{aligned}
\mathcal{L} = \alpha(\mathcal{L}_{CD}^{ref} + \mathcal{L}_{CD}^{r}) + \beta(\mathcal{L}_{CD}^{tar} + \mathcal{L}_{CD}^{p}) +  \gamma \mathcal{L}_{\mathcal{W}},
\end{aligned}
\end{equation}
where $\alpha$, $\beta$ and $\gamma$ are  weight factors. 
  
\subsection{Constraining the Shape Generation }\label{Shape constrain}
To provide further shape constraints, one can employ discriminators to assess both the latent features and point clouds, as done by the Latent.\cite{pc_latent} model. Hence, our framework can be further enhanced with such discriminators.   We call the variant of our framework that uses these discriminators \textit{RefComp w/dis}. Note that this variant is different from existing GAN-based solutions. Specifically, we do not use any GAN to learn distributions but only apply a discriminator to discriminate between the completed results and the reference data, which means that the training data can include several classes.

\section{Experiments}

\begin{table*}[!t]
\caption{Performance on the CRN dataset in terms of [CD↓ /F1↑], where CD is scaled by $10^4$.The \textbf{Best} and the \uline{Second-Best} performance are highlighted.} 

 \centering

 \scalebox{1}{ 
  \begin{tabular}{lccccccccc}

     \toprule Method & Plane  & Cabinet & Car & Chair & Lamp & Sofa & Table & Boat & Average\\\midrule
    
    FoldingNet~\cite{yang2018foldingnet}&2.4/--  & 8.4/-- & 4.9/-- &9.2/-- &11.5/-- &9.6/-- &8.4/-- &7.4/-- &7.7/-- \\
    
    PCN~\cite{yuan2018pcn}& 3.0/--  & 7.5/-- & 5.7/-- &9.7/-- &9.2/-- &9.5/-- &9.2/-- &6.2/-- &7.5/-- \\
    
    TopNet~\cite{sup_tchapmi2019topnet}& 2.3/--  & 8.2/-- & 4.7/-- &8.6/-- &11.0/-- &9.3/-- &7.5/-- &5.2/-- &6.4/-- \\
\midrule
   Pcl2pcl\cite{pc_unpaired_pcl2pcl} & 9.7/89.1  & 27.1/68.4 & 15.8/80.8 & 26.9/70.4 &25.7/70.4 &34.1/58.4 &23.6/79.0 &15.7/77.8 & 22.4/74.2\\

    Cycle.\cite{pc_cycle4completion} & 5.2/94.0& 14.7/\uline{82.1} & 12.4/82.1 & 18.0/77.5 & 17.3/77.4 & 21.0/75.2 & 18.9/81.2 & 11.5/84.8 & 14.9/81.8 \\

    Inversion.\cite{pc_shape_inversion}& 5.6/94.3 & 16.1/77.2 & 13.0/85.8 & 15.4/81.2 & 18.0/\textbf{81.7}&24.6/78.4 & 16.2/\textbf{85.5} &10.1/87.0 &14.9/83.9\\

    Latent. \cite{pc_latent}& 3.9/95.9  & \uline{13.5}/\textbf{83.3} & 8.7/\uline{90.4} & 13.9/82.3 &15.8/81.0 &14.8/\uline{81.6} &17.1/82.6 &10.0/87.6 &12.2/\textbf{85.6} \\

    P2C~\cite{cui2023p2c} & \textbf{3.5}/- & \textbf{11.7}/- & 9.0/- & 12.8/- & 16.4/- & 16.2/- & 18.6/- & \textbf{9.1}/- & 12.2/- \\

    R2C~\cite{zhang2024reverse2complete}& 4.7/- & -/- & -/- & 17.5/- & -/- & -/- & 18.8/- & -/- & -/-
   
    \\\midrule
    RefComp  & 3.8/\uline{96.5}  & 15.8/79.7 & \textbf{8.5}/\textbf{91.3} & \uline{12.4}/\uline{83.7} & 16.6/78.0 & \textbf{13.6}/\textbf{82.0} & \textbf{15.9}/\uline{84.0} & \uline{9.7}/\textbf{88.9} & \textbf{12.0}/\uline{85.5}
    \\

    RefComp w/dis &  \uline{3.6}/\textbf{96.6}  & 15.5/80.3 & \uline{8.7}/89.4 & \textbf{12.1}/\textbf{85.1} & \uline{15.1}/79.4 &\uline{14.6}/80.5 & \uline{16.0}/83.5 &10.3/\uline{88.2}  & \textbf{12.0}/85.4 \\ 
    \midrule
    
    RefComp Unified &  5.1/95.1  & 19.7/78.6 & 11.9/87.3 & 14.6/81.4 & \textbf{14.0}/\uline{81.6} &20.8/78.3 & 17.1/84.4 &10.8/87.8  & 14.3/84.3 \\\bottomrule
  \end{tabular}
 }

\label{result1}
\vspace{-0.5em}
\end{table*}

\begin{table*}[!t]
\caption{Performance on the 3D-EPN dataset in terms of [CD↓ /F1↑], where CD is scaled by $10^4$. The \textbf{Best} and the \uline{Second-Best} performance are highlighted.}
\centering
\label{3D-epn}
\begin{tabular}{lccccccccc}
\toprule
Method & Plane  & Cabinet & Car & Chair & Lamp & Sofa & Table & Boat & Average \\ \midrule

FoldingNet~\cite{yang2018foldingnet}& 2.6/--  & 7.6/-- & 4.8/-- &8.3/-- &9.7/-- &7.4/-- &8.0/-- &5.8/-- &6.8/-- \\

PCN~\cite{yuan2018pcn}& 2.5/--  & 8.0/-- & 4.8/-- &9.0/-- &12.2/-- &8.1/-- &8.9/-- &6.0/-- &7.4/-- \\

TopNet~\cite{sup_tchapmi2019topnet}& 2.3/--  & 7.5/-- & 4.6/-- &7.6/-- &8.9/-- &7.3/-- &7.5/-- &5.2/-- &6.4/-- \\

PoinTr~\cite{yu2021pointr}& 1.2/--  & 6.5/-- & 4.0/-- &5.1/-- &4.5/-- &5.4/-- &5.4/-- &2.6/-- &4.3/-- \\

\midrule
Pcl2pcl\cite{pc_unpaired_pcl2pcl} & 4.0/--  & 19.0/-- & 10.0/-- &20.0/-- &23.0/-- &26.0/-- &26.0/-- &11.0/-- & 17.4/-- \\

Cycle.\cite{pc_cycle4completion} & 3.7/96.4 & \uline{12.5}/\textbf{87.1} & 8.1/91.8 & 14.6/84.2 & 18.2/80.6 & 26.2/71.7 & 22.5/82.7 & 8.7/89.8 & 14.3/85.5 \\

Inversion.\cite{pc_shape_inversion}& 4.3/96.2 & 20.7/79.4 &11.9/86.0 & 20.6/81.1 & 25.9/78.4 & 54.8/74.7 &38.0/80.2 &12.8/85.2 &23.6/82.7\\

Latent. \cite{pc_latent}& 3.5/\uline{96.8}  &\textbf{12.2}/\uline{86.4} & 9.0/88.4 & 12.1/\textbf{86.4} &17.6/\textbf{81.6} &26.0/75.5 &19.8/85.5 &8.6/\textbf{89.8} &13.6/86.3\\

P2C*~\cite{cui2023p2c} & 3.7/-- & \uline{12.5}/-- & 7.7/-- & \textbf{11.3}/-- & \textbf{15.3}/-- & \textbf{13.2}/-- & 15.2/-- & \textbf{8.0}/-- & \textbf{10.9}/-- \\

R2C~\cite{zhang2024reverse2complete}& 4.5/-- & --/-- & --/-- & 14.0/-- & --/-- & --/-- & 12.5/-- & --/-- & --/-- \\

\midrule

RefComp & \uline{3.3}/97.3  & 13.9/83.2 & \textbf{7.3}/\uline{92.8} & 11.9/84.9 & 18.8/78.9 & 22.0/82.9 & 12.3/\uline{85.8}& 8.5/90.2 & 12.3/87.0 \\

RefComp w/dis &  \textbf{3.0}/\uline{97.7}  & 14.0/83.4 & \uline{7.4}/92.7 & \uline{11.5}/85.5 & \uline{16.5}/80.4 & 24.0/81.7 & \textbf{11.9}/\textbf{86.4}  &\textbf{8.0}/\uline{91.0}  & 12.0/87.4 \\ 

RefComp w/dis* &  \textbf{3.0}/\textbf{97.8}  & 13.9/83.3 & \uline{7.4}/\textbf{92.9} & \uline{11.5}/\uline{85.5} & 16.7/\uline{80.6} & 15.2/\uline{85.9} & \textbf{11.9}/\textbf{86.4}  & \textbf{8.0}/\textbf{91.1} & \textbf{10.9}/\textbf{87.9} \\
\midrule

RefComp Unified &  3.7/96.8 & 13.8/83.3 & \uline{7.4}/92.5 & 11.6/85.3 & 17.4/80.1 & 22.5/84.4 & 12.4/85.5 & \uline{8.1}/90.5 & 12.1/87.3  \\ 

RefComp Unified* &  3.7/96.9 & 13.8/83.5 & \textbf{7.3}/92.5 & 11.6/85.1 & 17.6/80.1 & \uline{14.7}/\textbf{85.9} & \uline{12.2}/85.5 & \uline{8.1}/90.2 &\uline{11.1}/\uline{87.5} \\
\bottomrule
\end{tabular}
\end{table*}

\begin{table}[!t]
\caption{Performance on the PartNet dataset in terms of MMD↓, where MMD is scaled by $10^2$.The \textbf{Best} and the \uline{Second-Best} performance are highlighted.}
 \centering
  \vspace{-0.8em}

\setlength{\tabcolsep}{0.8mm}{ 
 \scalebox{1}{ 
\begin{tabular}{lcccc}
\toprule \text { Method } & \text { Chair } & \text { Lamp } & \text { Table } & \text { Average } \\
\midrule \text { Pcl2pcl\cite{pc_unpaired_pcl2pcl} } & 1.90 & 2.50 & 1.90 & 2.10 \\
\text { MPC\cite{pc_mpc} } & 1.52 & 1.97 & 1.46 & 1.65 \\
\text { Cycle.\cite{pc_cycle4completion} } & 1.71 & 3.46 & 1.56 & 2.24 \\
\text { Inversion.\cite{pc_shape_inversion} } & 1.68 & 2.54 & 1.74 & 1.98 \\
\text { Latent.\cite{pc_latent} } & 1.43 & 1.95 & 1.37 & 1.58 \\
\midrule \text { RefComp } & \textbf{1.32} & 1.91 & 1.32 & 1.51 \\
\text {RefComp w/ dis.} & \uline{1.34} & \uline{1.78} & \textbf{1.30} & \uline{1.47} \\
\midrule \text {RefComp Unified} & 1.48 & \textbf{1.59} & \uline{1.31} & \textbf{1.46} \\
\bottomrule
\end{tabular}
}
 }
\vspace{1mm}

\label{result_mmd}
\vspace{-0.8em}
\end{table}

\subsection{Datasets}
The reference pairs are created from  ShapeNet\cite{chang2015shapenet}, which comprises over 50,000 point clouds spanning 55 object classes (see Section \ref{con_ref_data}). Before degrading the point clouds, we downsample the original point clouds to 2048 points to ensure having different sampling patterns.
To avoid reference partial point clouds from being too similar to the target partial point clouds, we use a minimum CD $ =1.0$ when searching for the $top-N$  most similar reference point clouds. The partial point cloud has 1024 points in each object, and the completed point cloud has 2048 points.
We use for evaluation the  virtual-scan datasets EPN-3D\cite{dai20173D-EPN}, CRN\cite{wang2020CRN}, PartNet\cite{mo2019partnet} with ground truth. These three datasets are derived from ShapeNet \cite{chang2015shapenet}, where the partial input of PartNet keeps the semantic part of each object. 
We follow the training-validation-testing split settings suggested by these datasets.
We also use the real-world dataset KITTI \cite{Geiger2013KITTI}, which is a benchmark dataset for autonomous driving. 
Following Inversion \cite{pc_shape_inversion}, we extract car objects from the KITTI dataset and align them with the data from ShapeNet regarding coordinate systems. Finally, we use the real-world datasets ScanNet \cite{dai2017scannet} and MatterPort3D \cite{chang2017matterport3d}, which consist of point cloud data acquired by depth cameras. 
It is worth noting that the MatterPort3D dataset contains only a small number of samples (20 for each class), and that none of the real-world datasets have complete point clouds as ground truth.

\subsection{Evaluation Metrics} 
We use the CD and F1-score to evaluate performance on the datasets with ground truth, as done by previous works\cite{pc_cycle4completion,pc_latent,pc_shape_inversion}. 
For the PartNet\cite{mo2019partnet} dataset, we follow \cite{pc_latent} and use the Minimum Matching Distance (MMD) as the metric. 
For the real-world datasets, we use the Unidirectional Chamfer Distance (UCD) and MMD. 
Details of these metrics are provided next:

\noindent \textbf{CD}: The Chamfer Distance measures the similarity between two point clouds (see Eq. \ref{Chamfer_distance}).

\noindent \textbf{F1-score}: It is defined as the harmonic average of the accuracy and the completeness. Accuracy measures the fraction of points in $\hat{c}_x$ that match the points in the ground truth $c_x$ 
and completeness measures the fraction of points in $c_x$ that match the points in  $\hat{c}_x$. 
In our experiments, we follow the setting used by Pcl2pcl \cite{pc_latent} with a threshold $\epsilon = 0.03$ for the $\mathcal{L}_2$ distance.

\noindent \textbf{MMD}:  It measures the fidelity between a set of completed point clouds $\mathbf{S}_c$ and a set of actual complete point clouds $\mathbf{S}_{gt}$. In this work, we  
compute the minimum distance between each sample in set $\mathbf{S}_c$ and samples in the ground truth set $\mathbf{S}_{gt}$, as done by \cite{pc_shape_inversion}. We then calculate the weighted average of these distances to obtain the MMD values.

\noindent \textbf{UCD}: It evaluates the
consistency from the partial point cloud $p_x$ to the completed point cloud $\hat{c}_x$, as follows:
\begin{equation}\label{ucd}
\begin{aligned}
{UCD} (p_x, \hat{c}_x) &=  \frac{1}{\vert p_x\vert}\sum_{a\in p_x}\min_{b\in \hat{c}_x}\Vert a-b\Vert_2^2,
\end{aligned}
\end{equation}
where $\{a, b\}$ are  points in  $p_x$ and $\hat{c}_x$, respectively.

\begin{figure*}[!t]
  \vspace{-0.6em}
	\centering
	\includegraphics[width=0.95\linewidth]{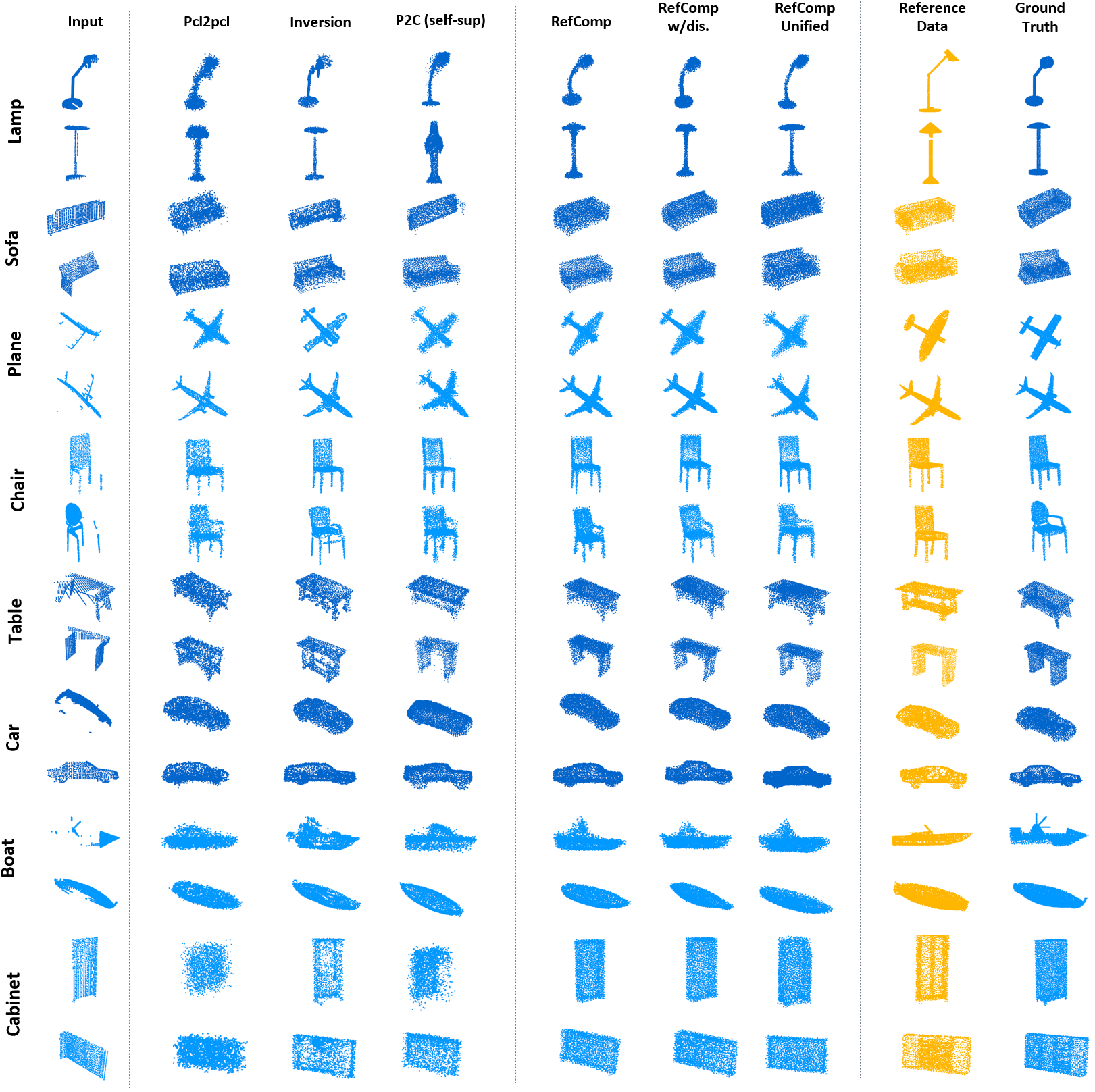}\
	\caption{\textbf{Sample completion results on the CRN dataset.} 
 All versions of our framework produce plausible results. The version  \textit{RefComp w/dis} further enhances edges.
 The reference data used by the three versions of \textit{RefComp} is shown in yellow. }
	\label{CRN_results}
  \vspace{-0.8em}
\end{figure*}

\subsection{Implementation Details}

To create the reference pairs $\{p_y, c_y\}$ based on a target partial input $p_x$ used as the template, we select the $top - k=15$ points in the degradation module. 
When we use the degradation module to obtain the partial point cloud, $p_{\hat{x}}$ (see Fig. \ref{mapping_jpg}), we empirically select the $top - k=5$ points.
In the LSFM, we expand the number of feature channels from 256 to 512 and reduce them back to 256 at the end. 
In the overall loss function in Eq. \ref{overall loss}, we set the weights $\alpha$, $\beta$ and $\gamma$ to 0.35, 0.65, and 0.001, respectively. 
For optimization, we use the AdamW optimizer with a learning rate of 5e-4, a weight decay of 5e-4, and a cosine scheduler. The batch size is set to 50, and the training is performed for 600 epochs. 
In the training process, we select the top three most similar reference pairs and randomly select one pair from these three at each iteration. 
During testing,
we only use the {reference pair} that is the closest to $p_x$ of the same class. 
All experiments are performed on 8 RTX 2080ti GPUs. 

Our framework can be trained in a class-aware setting, similar to previous unpaired point cloud completion models \cite{pc_cycle4completion, pc_latent,pc_shape_inversion}. In the following experiments, we evaluate the two versions of our framework (see Section \ref{Shape constrain}), i.e.,  \textit{RefComp} and \textit{RefComp w/dis} as trained under this class-aware setting. Let us recall that a key aspect of our framework
is its capability to generalize well when trained in a class-agnostic setting. 
When trained this way, our framework utilizes the same discriminators as those used by the  \textit{RefComp w/dis} version during the training. 
In the following experiments, we call this version of our model \textit{RefComp Unified}. Specifically, in \textit{RefComp w/ dis} and \textit{RefComp Unified}, we perform discrimination on $z_{c_{\hat{y}}}$ and $z_{c_{\hat{x}}}$ in the latent feature space $Z_{\mathcal{C}}$, and on $c_{\hat{x}}$ and $c_y$ in the point cloud space $\mathcal{C}$.

\subsection{Results on Virtual-scan Datasets}\label{vd}
We first evaluate the performance of our framework and other methods on the virtual-scan datasets CRN, 3D-EPN, and PartNet. 

\noindent \textbf{CRN  dataset.} \label{quant virtual}
Overall, the two class-aware versions of our framework perform very competitively on the CRN dataset, as tabulated in Table \ref{result1}. 
Compared with  Latent.\cite{pc_latent}, our framework attains better performance for several classes, regardless of the use of discriminators. 
Specifically, for the classes ``chair'', ``table'', and ``lamp'', whose objects have fine details,  e.g., legs of chairs and tables, and the neck of lamps, \textit{RefComp w/dis}  surpasses Latent.\cite{pc_latent} by 1.8, 1.1 and 0.7, respectively, in terms of the CD. On average over the eight classes, the two class-aware versions of our framework achieve similar performance, which is close to that achieved by Latent.\cite{pc_latent}. 
Regardless of the class, 
our framework can introduce strong structural features to provide shape constraints. It is worth mentioning that the additional discriminators in \textit{RefComp w/dis}  can enhance the feature fusion for the fine-detailed classes but may not improve performance for other classes in terms of the CD/F1-score metrics.
Compared to P2C~\cite{cui2023p2c}, our framework demonstrates superior overall performance in the class-aware setting. Furthermore, compared to R2C~\cite{zhang2024reverse2complete}, our framework achieves better performance on the `plane', `chair', and `table' classes.

\textit{RefComp Unified}'s performance on the CRN dataset is not as strong as that of the other two versions of our framework. 
However, this version still outperforms Cycle.\cite{pc_cycle4completion} and Inversion.\cite{pc_shape_inversion}, which rely on class-aware training. 
These results demonstrate the effectiveness of our framework as a class-agnostic solution.
For the ``lamp'' class, \textit{RefComp Unified} surpasses all other methods in terms of the  CD and ranks second best in terms of the F1-score.
We argue that
the class-agnostic training setting allows our model to learn more structural features from other classes that are useful for the ``lamp’’ class.
For other classes, e.g.,   ``cabinet'' and ``sofa'', \textit{RefComp Unified} shows an important decrease in performance compared to the other two versions of our framework. Specifically, for these two classes, the  CD drops by 4.2 and 6.2, respectively, compared to the \textit{RefComp w/dis}.
This may be because dense objects in these two classes have points distributed not only on the surface but also inside, making the completion task more challenging.

Note that for classes whose objects have different sizes and the partial point clouds depict very distinct sections of the object, 
 e.g., the ``cabinet'' class, the reference data may not provide accurate structural features for completion. This is the main reason for our framework's performance in the ``cabinet'' class, which is also observed in the ``cabinet'' class of the 3D-EPN dataset.

\noindent \textbf{3D-EPN dataset.} 
\textit{RefComp w/dis} attains the best performance on average over all classes of this dataset, as tabulated in Table \ref{3D-epn}. 
The two class-aware versions of our framework substantially outperform the SOTA on the ``car'', ``sofa'', and ``table'' classes. 
We attribute these results to the utilization of reference data.
Furthermore, it is worth mentioning that the \textit{RefComp Unified} attains more stable results
on the 3D-EPN dataset compared to those attained on the CRN dataset. 
As shown in Table \ref{3D-epn}, \textit{RefComp Unified}'s performance is, on average, better than that of \textit{RefComp}, and only slightly below than that of
\textit{RefComp w/dis}. 
Moreover, \textit{RefComp Unified} outperforms all other class-aware models. 
We argue that a class-agnostic training setting allows the models to learn from a wide range of structural features from all classes, which enhances performance. 

Following P2C in \cite{cui2023p2c}, we follow their approach by replacing the ``sofa'' category with the same dataset used in \cite{cui2023p2c}. We conduct additional experiments, whose results are tabulated in  Table~\ref{3D-epn} as P2C*, RefComp w/dis* and RefComp Unified*.
While maintaining stable performance in other categories, we observe significant improvements in the ``sofa'' category, with performance increasing from 24.0 to 15.2 and from 22.5 to 14.7, respectively. 
Additionally, the average performance in both class-aware and class-agnostic settings reaches 10.9 and 11.1, respectively. Notably, while the class-aware setting achieves comparable performance to {P2C*}, the class-agnostic setting still delivers acceptable performance. These results further validate the effectiveness of our framework in class-agnostic scenarios.

Note that the partial point clouds in the 3D-EPN dataset are not as sparse as those in the
CRN dataset and the distribution of the complete point clouds along object surfaces
is not as uniform as that found in the complete point clouds of the CRN dataset. These characteristics lead to facilitate the
completion task in this dataset. 

\noindent \textbf{PartNet dataset.} 
The partial point clouds of the PartNet depict only a few semantic parts of the complete objects, making the completion task more challenging. 
Table \ref{result_mmd} tabulates results for this dataset in terms of the MMD. 
The two class-aware versions of our framework attain SOTA results for all classes. \textit{RefComp w/dis} even surpasses \textit{RefComp} by 0.13 MMD for the ``lamp'' class.
Although \textit{RefComp Unified} attains a lower performance than that of 
\textit{RefComp w/dis}
on the ``chair'' class, its performance on the ``lamp'' class is the best with 1.58 MMD,  surpassing by  0.19 MMD the second-best performing model.
\textit{RefComp Unified} outperforms, on average, 
all class-aware models.
These results show that our framework works well not only as a class-aware solution but also as a class-agnostic solution.

\begin{table}[!t]
\caption{Performance on real-world datasets in terms of  UCD ↓, where UCD is scaled by $10^4$.The \textbf{Best} and the \uline{Second-Best} performance are highlighted.}
\renewcommand\arraystretch{1}
 \centering
  \vspace{-0.2em}

\setlength{\tabcolsep}{0.5mm}{ 
 \scalebox{1}{ 
 \begin{tabular}{lccccc}
\toprule  Method   & \multicolumn{2}{c}{\text { ScanNet }} & \multicolumn{2}{c}{\text { MatterPort3D }} & \text { KITTI } \\
\cmidrule(l){2-6} &  Chair  & Table  & Chair  & Table & Car \\
\midrule 
\text { Pcl2pcl \cite{pc_unpaired_pcl2pcl} }  & 17.3 & 9.1 & 15.9 & 6.0 & 9.2 \\
\text { Cycle. \cite{pc_cycle4completion} }  & 9.4 & 4.3 & 4.9 & 4.9 & 9.4 \\
\text { Inversion. \cite{pc_shape_inversion}}  & 3.2 & 3.3 & 3.6 & 3.1 & 2.9 \\
\text { Latent. \cite{pc_latent}}  & 3.2 & 2.7 & 3.3 & 2.7 & 4.2 \\
\text { Energy. \cite{cui2022energy}}  & 4.0 & 3.0 & 4.0 & 3.0 & 7.3 \\
\midrule
\text { RefComp } & 3.2 & 2.5 & 3.3 & \uline{2.5} & 3.1 \\
\text { RefComp w/dis.} & \uline{3.1} & \uline{2.4} & \uline{3.2} & 2.7 & \uline{3.0} \\
\midrule
\text { RefComp Unified } & \textbf{2.8} & \textbf{2.2} & \textbf{2.8} & \textbf{2.1} & \textbf{2.7} \\
\bottomrule
\end{tabular}
}
 }
 \vspace{1mm}
\label{result_ucd}
\vspace{-0.8em}
\end{table}

\noindent \textbf{Qualitative results.} 
In Fig. \ref{CRN_results}, we show several completion results of the different versions of our framework for all the classes in the CRN dataset.
We compare these results with those of  Pcl2pcl \cite{pc_unpaired_pcl2pcl} and Inversion \cite{pc_shape_inversion}. The last column of 
 Fig. \ref{CRN_results} also shows the reference data used as input by our three versions of the \textit{RefComp} framework (yellow column).  
Let us recall that CRN is a virtual-scan dataset, hence we can use the ground truth to calculate appropriate metrics for qualitative comparisons. However, it is important to keep in mind that our

Note that the results of Pcl2pcl \cite{pc_unpaired_pcl2pcl} exhibit strong dissimilarities with the ground truth, particularly outliers for the ``lamp'' and ``cabinet'' classes.
Although the visual results for the remaining classes are strong, overall, the completed shapes do not show a strong correlation with the shape that one can infer from the partial input. 
One main issue in the visual results of  Inversion \cite{pc_shape_inversion} 
is the extremely non-uniform point distribution, especially for the ``table'' class, where the completed point clouds are not continuous in space. This method performs particularly unsatisfactorily when the partial input is very sparse, and hence, strong semantic features cannot be extracted,  e.g., the ``car'' and ``couch'' classes. 

In the case of the two class-aware versions of our framework, it can be observed that the presence of a discriminator has no significant impact on the qualitative results. Both versions exhibit similar performance, which is also consistent with the quantitative performance. 
One minor difference between these two versions, however, is the smoother edges that
\textit{RefComp w/dis}  completes compared to those completed by
\textit{RefComp}.
Overall, compared with other methods, our class-aware frameworks attain more plausible results. This shows that our framework can benefit from the structural features extracted from the reference data as well as those available in the partial input. 
For example, for the second instance of the ``lamp'' class, our framework learns to complete the lamp base from the reference data and fuses this information into the partial input. 
For the first instance of the ``table'' class, the reference data provides strong structural features to complete the partial input while maintaining the overall shape of that input.
We also evaluate the self-supervised version of P2C, and as shown in Fig.~\ref{CRN_results}, its qualitative results are significantly inferior than those of  RefComp. Notably, there are obvious errors in the completion of the ```lamp'' and ``car'' categories, and the results for the ``couch'' and ``cabinet'' classes fall short of expectations.

The class-agnostic version of our framework, i.e.,  \textit{RefComp Unified}, overall,  shows consistent qualitative performance. Although this version slightly 
decreases the smoothness of the completed results compared to the class-ware versions, it still produces plausible shapes, e.g., see the results for the ``plane,'' ``car,'' ``lamp,'' and ``chair'' classes. 
Note that the results for the ``sofa" and ``cabinet" classes exhibit the most significant decrease in quality compared to those attained by the class-aware versions. These results are consistent with
the quantitative performance in terms of the CD/F1-score metrics.
In this class-agnostic version of our framework, the reference data can be considered as `pseudo labels,’   
as the class labels are not accurate. This makes it more challenging 
to effectively complete sparse and ambiguous partial point clouds like those in the ``sofa" and ``cabinet" classes. 
\begin{figure*}[!t]
  \vspace{0em}
	\centering
	\includegraphics[width=0.8\linewidth]{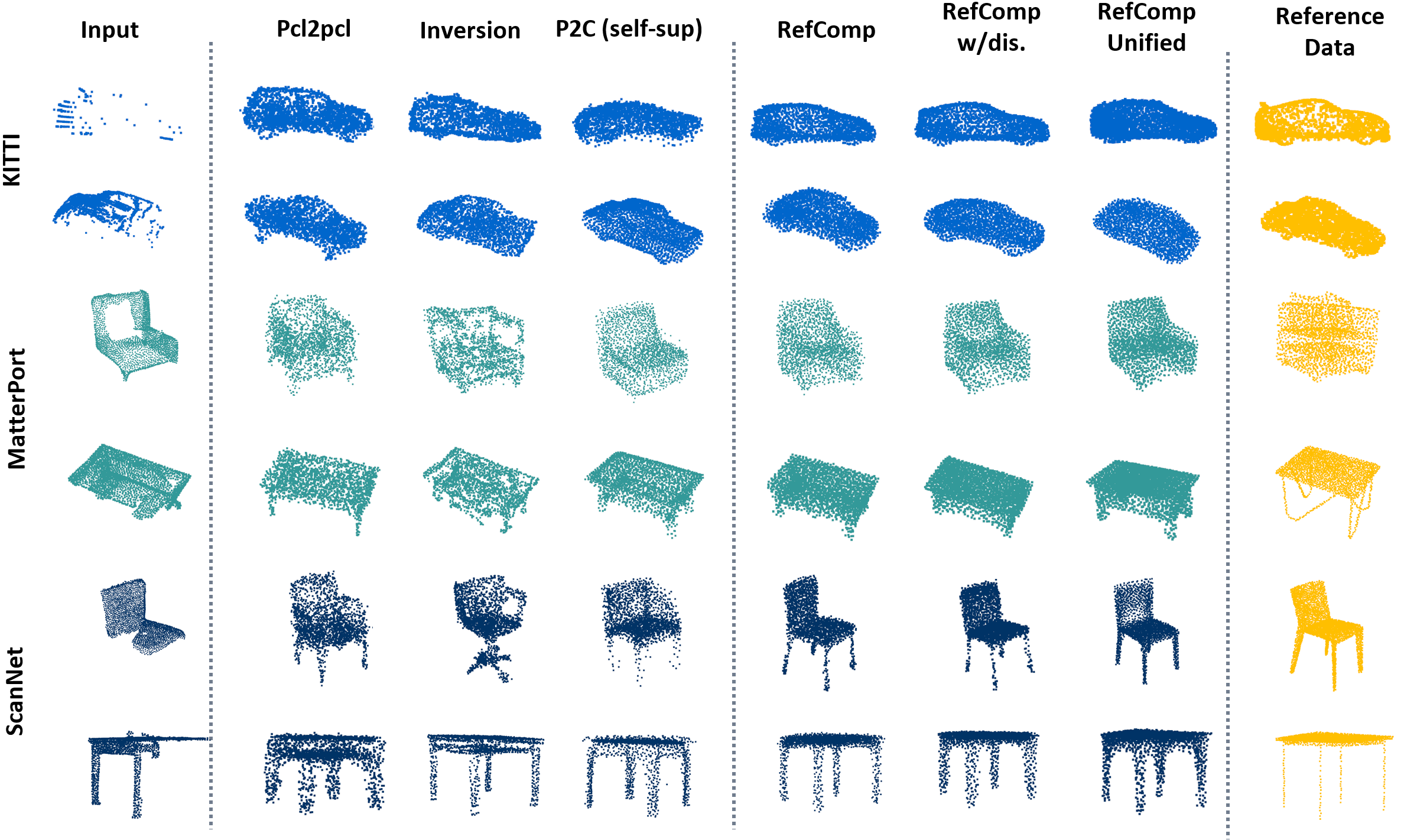}\
	\caption{\textbf{Sample completion results on real-world datasets.}  
 Our framework, with or without discriminators, produces more plausible results than other methods. 
 The reference data used by three versions of \textit{RefComp} is shown in yellow.
 }
	\label{real_world}
  \vspace{-0.8em}
\end{figure*}

\begin{table}[!t]
\caption{Performance on real-world datasets in terms of MMD ↓. The \textbf{Best} and the \uline{Second-Best} performance are highlighted.}
 \centering
  \vspace{-0.2em}

\setlength{\tabcolsep}{1.8mm}{ 
 \scalebox{1}{ 
\begin{tabular}{l|c|ccccc}
\toprule   
\multirow{2}{1cm}{Methods}   &  \multirow{2}{0.6cm}{\textit{sup.}}  & \multicolumn{2}{c}{ ScanNet } & \multicolumn{2}{c}{ MatterPort3D } & KITTI \\
\cmidrule(l){3-7} & $ $ & Chair  & Table  & Chair  & Table & Car \\
\midrule 
GRNet \cite{xie2020grnet} & sup. & 6.070 & 6.302 & 6.147 & 6.911 & 2.845 \\
PoinTr \cite{yu2021pointr} & sup. & 6.001 & 6.089 & 6.248 & 6.648 & 2.790 \\
\midrule
Cycle. \cite{pc_cycle4completion}& unpaired & 6.278 & 5.727 & 6.022 & 6.535 & 3.033 \\
Inversion. \cite{pc_shape_inversion} & unpaired & 6.370 & 6.222 & 6.360 & 7.110 & 2.850 \\
Latent. \cite{pc_latent}& unpaired & 5.893 & 5.541 & 5.770 & 6.076 & 2.742 \\
\midrule  
RefComp & unpaired & \uline{5.645} & \textbf{3.633} & \uline{5.584} & \uline{3.787} & \uline{2.685} \\
RefComp w/dis.  & unpaired & \textbf{5.342} & \uline{3.771} & \textbf{5.576} & \textbf{3.750} & \textbf{2.532} \\
\midrule  
RefComp Unified  & unpaired & 6.144 & 4.157 & 6.213 & 4.195 & 2.804 \\
\bottomrule
\end{tabular}
}
 }
 \vspace{1mm}
\label{result_mmd_real}
\vspace{-1em}
\end{table}

\subsection{Results on Real-world Datasets}
In this section, we evaluate our framework
on three indoor and outdoor real-world datasets, KITTI, ScanNet, and MatterPort3D, following \cite{pc_latent, pc_shape_inversion}, to further confirm our framework's
quantitative and qualitative performance. 
Note that these datasets are sparser and lack semantic features compared to the virtual-scan datasets, leading to more challenges. 

As tabulated in Table \ref{result_ucd}, the two class-aware versions of our framework
attain similar results to those attained by other unpaired point cloud completion models. However, the class-agnostic version, i.e.,  \textit{RefComp Unified}, achieves  SOTA results on different classes and datasets, surpassing the class-aware versions by up to 0.6 UCD. 
This shows the generalization capabilities of \textit{RefComp Unified} and confirms the advantages of using reference data.

Table \ref{result_mmd_real} tabulates the fidelity between a set of complete and completed point clouds in terms of the MMD. 
To further confirm the strong performance of our framework, this table includes 
two advanced supervised completion methods, GRNet \cite{xie2020grnet} and PoinTr \cite{yu2021pointr}, which have been fine-tuned to these datasets.  
It is clear that the two class-aware versions of our framework, without any fine-tuning, attain stronger performance, surpassing even the
the supervised methods. 
For the ``table'' class,  \textit{RefComp} surpasses Latent. \cite{pc_latent}  by 1.908 MMD and 2.289 MMD on the ScanNet and MatterPort3D datasets, respectively. 
These results indicate that the two class-aware versions of our framework effectively preserve the semantic features of each class. 
The qualitative results in Fig. \ref{real_world} show that \textit{RefComp} and \textit{RefComp w/dis} produce more stable and plausible completion results, while those attained by
Inversion.\cite{pc_shape_inversion} appear sparse. 
Note that when the shape of the reference and the partial input are not sufficiently similar, 
the completed shapes tend to resemble the partial input. 
For instance, this can be observed in the  ``chair'' and ``table'' classes of the MatterPort dataset. 
This confirms our framework's robustness to a lack of sufficiently similar references for a given partial input.

The performance of  \textit{RefComp Unified} is consistent and competitive with an expected decrease in MMD values compared to \textit{RefComp} and \textit{RefComp w/dis} (see Table \ref{result_mmd_real}). However, the visual results of \textit{RefComp Unified} as shown in Fig. \ref{real_world} are evidently inferior to those of \textit{RefComp} and \textit{RefComp w/dis}. 
For instance, see the second instance of the  ``car'' class (KITTI dataset). 
As discussed before,  each object does not have an explicit label in a class-agnostic training setting and the reference data only provides a `pseudo label,' which prevents \textit{RefComp Unified} from completing all partial objects effectively.

 \begin{figure*}[!t]
  \vspace{-0.2em}
	\centering
	\includegraphics[width=1\linewidth]{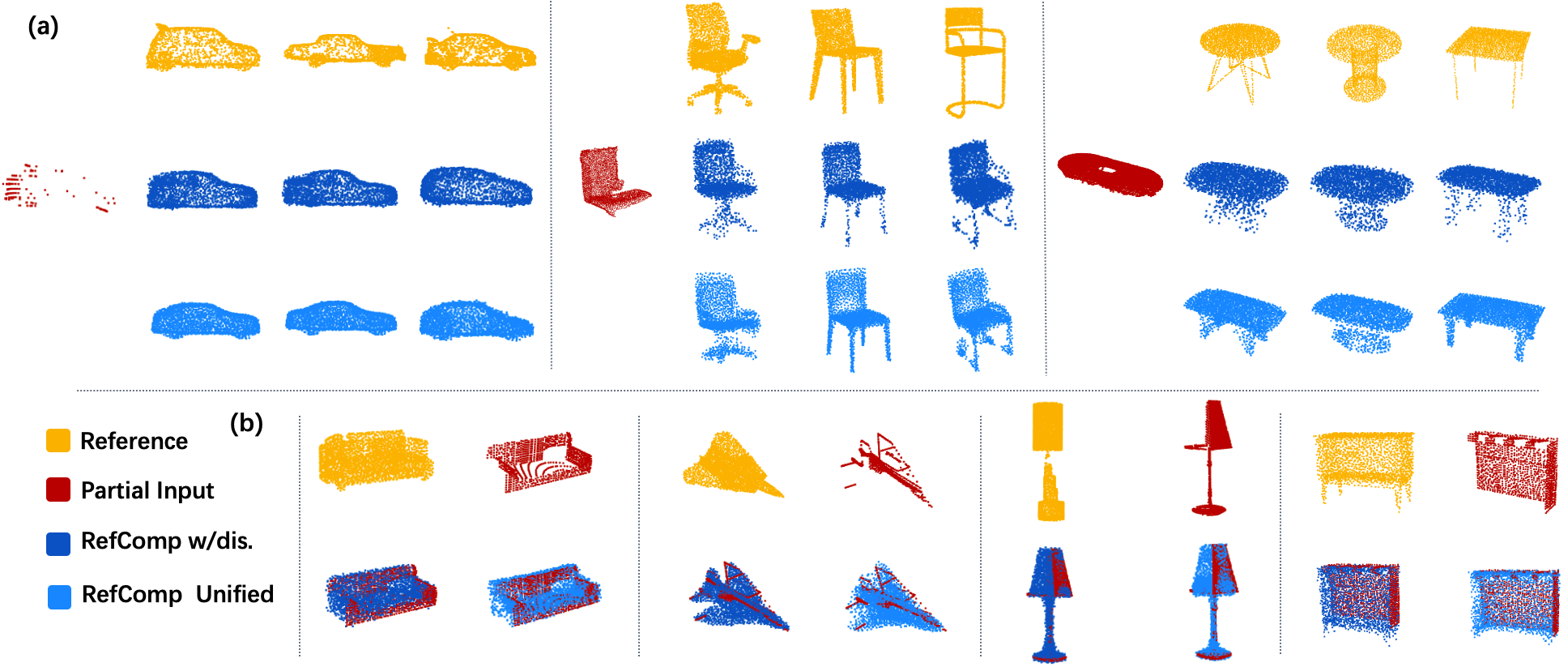}\
	\caption{\textbf{Additional experiments.} 
    (a) Completion results with different reference data on real-world datasets (left to right) KITTI, ScanNet, and MatterPort.
    (b) Overlapping the partial input on the completed results on the CRN dataset to demonstrate the framework's capability of preserving the overall shape 
    (best view in color.) 
 }
\label{more_exp}
  \vspace{-0.2em}
\end{figure*}

\subsection{Additional Experiments}
We conduct additional experiments to further demonstrate our model's effectiveness.
We first compute qualitative results with different reference data on real-world datasets. 
We then show qualitative results by overlapping the partial input on the completed results to showcase our framework's capability of preserving the overall shape.
Due to the similar performance of \textit{RefComp} and \textit{RefComp w/dis}, we only use \textit{RefComp w/dis} in these additional experiments.

\noindent \textbf{Results with different reference data.} 
When the partial input is extremely sparse, different reference data can provide distinct information to achieve the completion. In Fig. \ref{more_exp}(a), we show three very sparse partial point clouds and the corresponding results when a different reference is used. It can be observed that different reference data guide the completion process differently, regardless of using a class-aware or class-agnostic training setting.
While the results of \textit{RefComp w/dis} exhibit relatively rough edges, they retain the overall shape of the partial input. 
On the other hand, the results of \textit{RefComp Unified} have smoother edges, but the preservation of the overall shape is not as clear as in the case of \textit{RefComp w/dis}.

\noindent \textbf{Overlapping results.} 
In Fig. \ref{more_exp}(b), we show results by overlapping the partial input on the completed point cloud.
Although \textit{RefComp Unified} does not tend to preserve the overall shape as \textit{RefComp w/dis} does, e.g., see the ``plane'' class, these results show, overall,
the plausibility of the shapes completed by our framework, regardless of the training setting used. 

\begin{table}[!t]
  
\caption{Model size and performance of our pipeline on the KITTI dataset. UCD↓  values are scaled by $10^4$.}
\renewcommand\arraystretch{1}
\centering

\scalebox{0.80}{ 
\begin{tabular}{lccccc}
\toprule Methods &Params.& \textit{sup.}&\textit{cls-aw./ag.}& UCD & Infer. time(s)  \\
\midrule PoinTr \cite{yu2021pointr} &30.9M& sup. & cls-ag & 1.9 & 0.04 \\
 Inversion \cite{pc_shape_inversion} & 40M&unpaired & cls-aw & 2.9 & 30 \\ 
 P2C \cite{cui2023p2c} &23.1M & self-sup. & cls-aw & 1.8 & 0.03 \\
\midrule Generalized + Random &40.5M &unpaired& cls-ag & 2.7 &  0.04 \\
Generalized + Matching & 40.5M&unpaired& cls-ag& 2.7 & 0.02  \\
Fine-tuned + Random &40.5M &unpaired& cls-ag & 1.5 &  0.04 \\
Fine-tuned+ Matching &40.5M &unpaired& cls-ag & \textbf{1.3} & 0.02  \\
\bottomrule
\end{tabular}
}
\label{real_world_inference}
\vspace{-0.8em}
\end{table}

\noindent \textbf{Model size and extra real-world experiments on the KITTI dataset.} 
To validate the feasibility of deploying our point cloud completion pipeline in real-world scenarios, we conduct additional experiments on the KITTI dataset.
As shown in Table~\ref{real_world_inference}, our class-agnostic unpaired model is comparable in model size to supervised methods like PoinTr, unpaired methods like Inversion, and self-supervised model P2C, all within the same order of magnitude. Since our method requires retrieving similar reference point clouds to guide the completion process, we implement two strategies: one involves randomly retrieving point clouds for completion, and the other uses pre-retrieved reference point clouds.
We evaluate all these methods on the same testing platform, measuring the completion time for 50 objects and using the average time. Additionally, we test the generalization performance of models trained on the 3D-EPN dataset and further fine-tuned on the KITTI dataset. As shown in Table~\ref{real_world_inference}, our approach achieves the best completion results after fine-tuning. Although the random search strategy is slower than the matching-based strategy,  with 0.04s per/object and 0.02s per/object, respectively,  it remains within an acceptable range. These findings demonstrate the high feasibility of integrating our point cloud completion pipeline into autonomous driving frameworks.

\begin{table}[t]
\vspace{-1.5em}
\caption{Performance on the CRN dataset for several ablation experiments in terms of CD↓, where CD↓ is scaled by $10^4$. } 
\centering

\scalebox{1}{ 
\begin{tabular}{lccccc}
\toprule Method & Chair & Lamp & Sofa & Table & Avg. \\
\midrule 
{RefComp w/dis} & \textbf{12.8} & \textbf{16.6} & \textbf{13.6} & \textbf{15.9} & $\textbf{14.4}$ \\
Fixed Ref. & 15.2 & 18.2 & 14.9 & 17.4 & 16.4 \\
Cross-T & 19.2 & 19.5 & 15.3 & 17.2 & 17.8 \\
Only GAN & 13.4 & 17.8 & 14.1 & 18.1 & 15.9 \\
No Share & 36.8 & 26.2 & 30.5 & 30.8 & 31.1 \\

 \midrule
{RefComp unified} & \uwave{14.6} & \uwave{14.0} & \uwave{20.8} & \uwave{17.1} & \uwave{16.6} \\
Unified Fixed Ref. & 14.8 & 16.1 & 23.1 & 17.8 & 18.0 \\
Unified Cross-Attn & 15.5 & 16.4 & 21.9 & 17.2 & 17.8 \\
Unified Only GAN & 25.4 & 20.0 & 37.6 & 20.1 & 25.8 \\
Unified No Share & 29.6 & 33.7 & 34.5 & 25.4 & 30.8 \\

\bottomrule
\end{tabular}
 }

\label{ablation_all}
\vspace{-1.5em}
\end{table}

\subsection{Ablation Studies}

To verify the effectiveness of each component of the proposed
framework, we conduct three different ablation experiments on four classes of the CRN dataset. We use the versions 
\textit{RefComp w/dis} and 
\textit{RefComp Unified} in these experiments. 
All results are tabulated in Table \ref{ablation_all}.

\noindent \textbf{Fixed reference.} To confirm the advantages of randomly selecting one reference pair from the $top-3$  at each training iteration, 
we use one fixed reference pair in this experiment, i.e., the most similar reference pair.
The results are shown in Table \ref{ablation_all} as Fixed Ref. and Unified Fixed Ref. for the two versions of our framework, respectively.
Averaged over the four classes, the performance of \textit{RefComp w/dis} drops from 14.4 to 16.4 CD, and that of \textit{RefComp Unified} drops from 16.6 to 18.0 CD. This proves that randomly selecting reference data within the set of $top-3$ reference pairs enables the framework to learn more structural information.

\noindent \textbf{Replacement and removal of LSFM.}
We replace the LSFM with a Cross-Attention, which follows the transformer architecture of \cite{chen2021crossvit}.
We also remove the LSFM and only use a GAN with a discriminator to complete the partial point clouds.
The GAN's architecture follows the works in \cite{pc_cycle4completion} and \cite{pc_latent}.
The results of this study are tabulated in Table \ref{ablation_all} as Cross-Attn and Only GAN, and Unified Cross-Attn and Unified Only GAN, for the two versions of our framework, respectively.  
Note that although the Cross-Attn versions are expected to perform strongly, their performance is not the best in these experiments. 
We argue that in the 3D space, the distance between constituent parts of the target point cloud as computed by the transformer, may not directly reflect their semantic relevance, unlike in images.
Compared to using the LSFM, the low CD values attained by the Only GAN versions confirm the advantages of using our feature fusion 
to improve completion performance, i.e., by enhancing latent features and enabling the framework to learn more structural information.

\noindent \textbf{Parameter-sharing mechanism}. 
In this ablation experiment, we remove the parameter sharing mechanism between the reference branch and the target branch. 
We train this version from scratch.
The results are shown in Table \ref{ablation_all} as No Share and Unified No Share for the two versions of our framework, respectively.
As observed, without the parameter sharing mechanism, the CD metric for both versions of our framework sees a significant reduction. 
This further substantiates the key role of the parameter sharing mechanism in our framework.

\section{Conclusion and Future Work}
In this paper, we introduced \emph{RefComp}, a unified framework for unpaired point cloud completion that can work as a class-aware or class-agnostic solution. 
Distinct from previous methods, \emph{RefComp} uses reference data as guidance, which provides a distribution that is similar to that of the partial point clouds to be completed. 
Using this reference data, \emph{RefComp}  transforms the unpaired point cloud completion problem into a \emph{shape translation} problem in the latent feature space. 
\emph{RefComp} uses a reference branch and a target branch to complete the reference partial data and the target partial data, respectively, with a parameter-sharing mechanism.
The \emph{shape translation} problem is solved by a novel Latent Shape Fusion Module (LSFM) that fuses structural features from reference data to the target partial point cloud data. 
The LSFM then allows 
achieving both plausible and stable results.
Under a class-aware training setting,  \emph{RefComp}  achieved SOTA results for several classes while achieving a very competitive performance under a 
class-agnostic training setting. 
For future work, we aim to better integrate point cloud completion with downstream tasks. For example, we plan to enhance the completion of indoor scenes captured by depth cameras, which can significantly improve the realism of scene reconstruction. Additionally, we hope to combine point cloud completion with robotic applications, enabling robots to better perceive 3D spaces and generate more reasonable grasping poses using the completed 3D point clouds.

\bibliographystyle{IEEEtran}
\bibliography{reference}

\begin{IEEEbiography}
[{\includegraphics[width=1in,height=1.25in,clip,keepaspectratio]{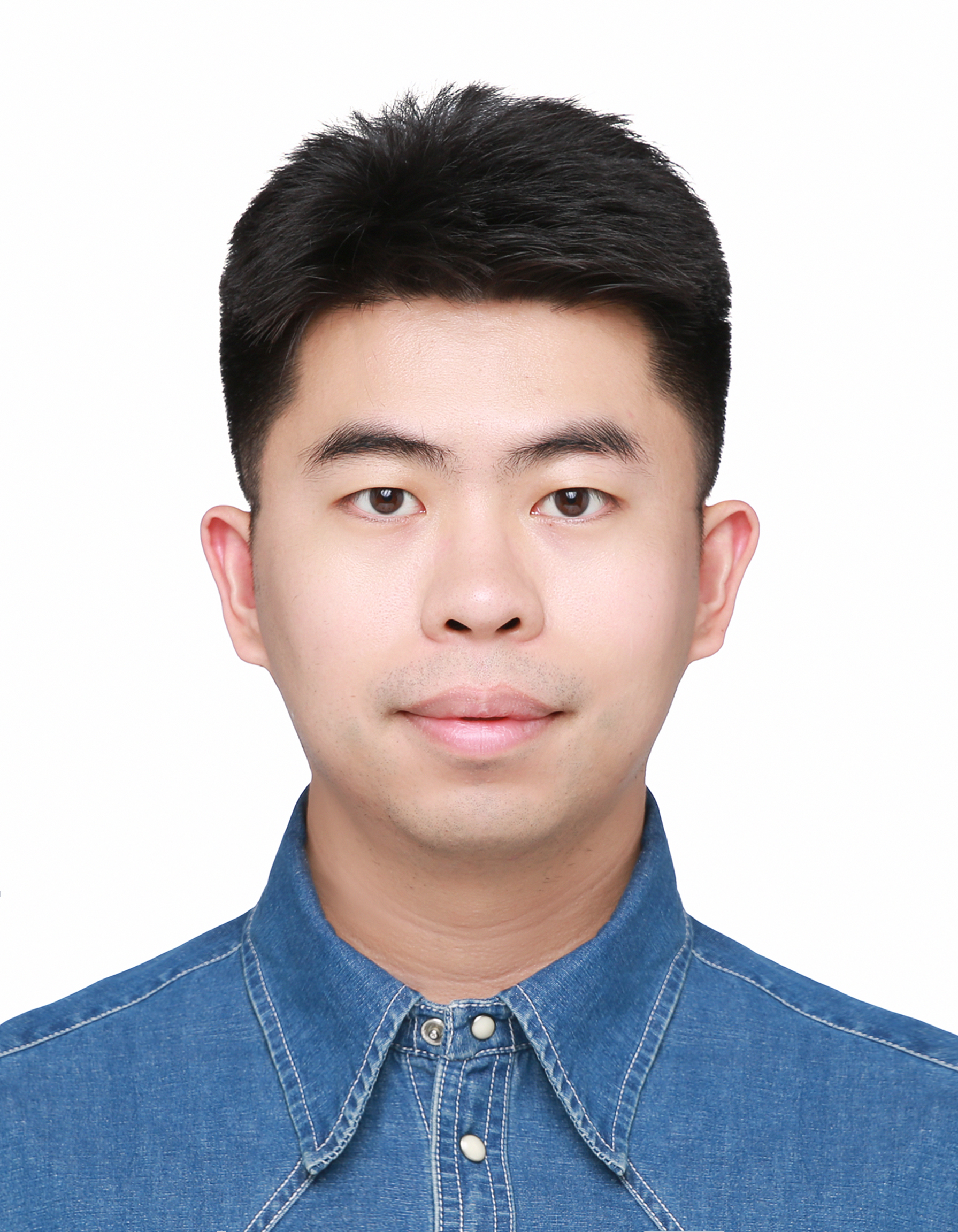}}]{Yixuan Yang} is pursuing a Ph.D. degree in Computer Science and Engineering, the University of Warwick, Coventry, UK and Computer Science and Engineering, the Southern University of Science and Technology, Shenzhen, China. He received an M.S. degree in Computer Science and Engineering from The Chinese University of Hong Kong, Hong Kong, in 2020. His research interests include 3D computer vision, point cloud, and embodied AI.
\end{IEEEbiography}
\begin{IEEEbiography}
[{\includegraphics[width=1in,height=1.25in,clip,keepaspectratio]{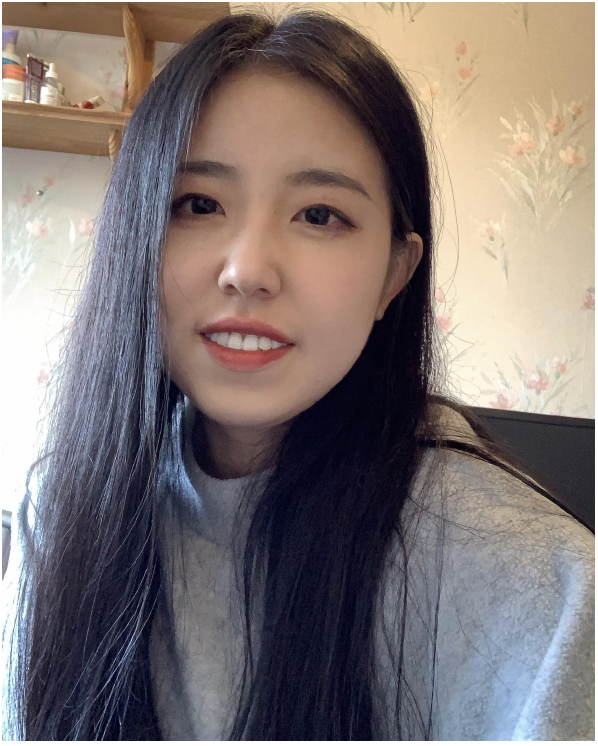}}]{Jinyu Yang} received a Bachelor's degree in Engineering from Beihang University, Beijing, China, in 2018, an M.S. degree in Electrical and Electronic Engineering from The Hong Kong University of Science and Technology, Hong Kong, in 2019, and a Ph.D. degree from the University of Birmingham and Southern University of Science and Technology (SUSTech) in 2024. She is currently with Tapall.ai, Shenzhen, China. Her research interests include computer vision and object tracking.
\end{IEEEbiography}
\begin{IEEEbiography}
[{\includegraphics[width=1in,height=1.25in,clip,keepaspectratio]{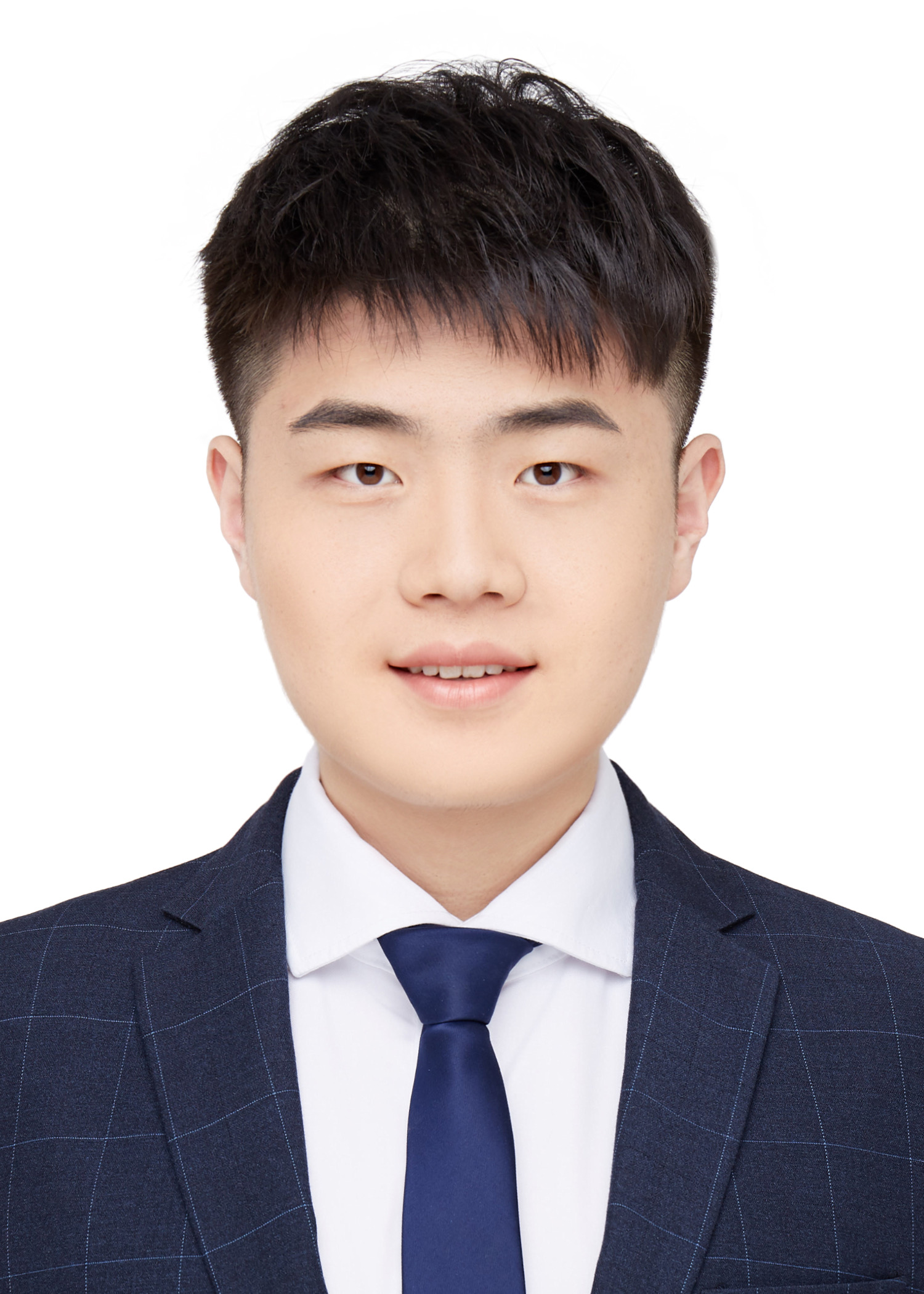}}]{Zixiang Zhao}
is currently a postdoctoral researcher at the Photogrammetry and Remote Sensing Group, ETH Zürich, Switzerland. He received his Ph.D. degree in statistics from the School of Mathematics and Statistics, Xi’an Jiaotong University, Xi’an, China. Previously, he was a visiting Ph.D. student at the Computer Vision Lab, ETH Zürich, Switzerland, and also worked as a research assistant at the Visual Computing Group, Harvard University, USA. His research interests include computer vision, machine learning, deep learning, with a particular focus on image/video restoration, low-level vision, and computational imaging.
\end{IEEEbiography}
\begin{IEEEbiography}
[{\includegraphics[width=1in,height=1.25in,clip,keepaspectratio]{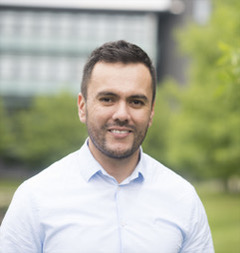}}] {Victor Sanchez} (Member, IEEE) received the M.Sc. degree from the University of Alberta, Canada, in 2003, and the Ph.D. degree from The University of British Columbia, Canada, in 2010. From 2011 to 2012, he was with the Video and Image Processing Laboratory, University of California at Berkeley, as a Postdoctoral Researcher. In 2012, he was a Visiting Lecturer with the Group on Interactive Coding of Images, Universitat Autònoma de Barcelona. From 2018 to 2019, he was a Visiting Scholar with the School of Electrical and Information Engineering, The University of Sydney, Australia. He is currently a Professor with the Department of Computer Science, University of Warwick, U.K. His main research interests are in the area of signal and information processing with applications to multimedia analysis, image and video coding, security, and communications. He has authored several technical papers in these areas and co-authored a book (Springer, 2012). His research has been funded by the Consejo Nacional de Ciencia y Tecnologia, Mexico; the Natural Sciences and Engineering Research Council of Canada; the Canadian Institutes of Health Research; the FP7 and H2020 programs of the European Union; the Engineering and Physical Sciences Research Council, U.K.; and the Defence and Security Accelerator, U.K.
\end{IEEEbiography}
\begin{IEEEbiography}
[{\includegraphics[width=1in,height=1.25in,clip,keepaspectratio]{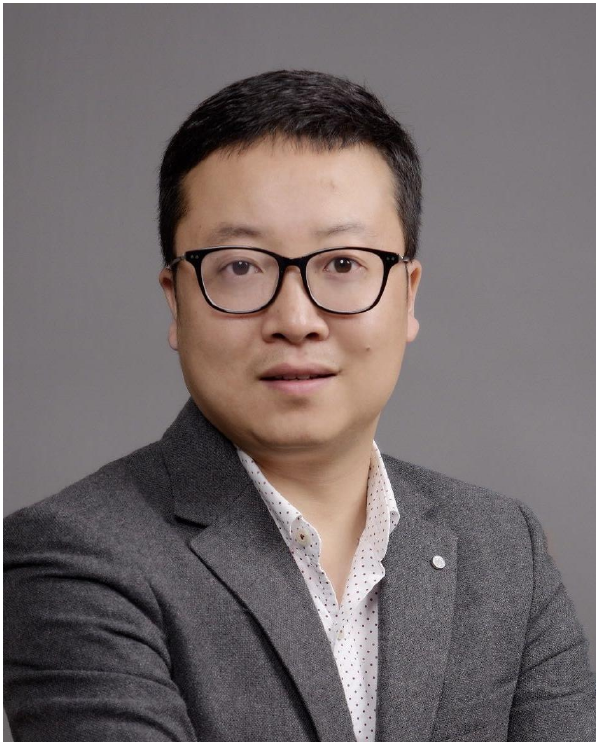}}]{Feng Zheng} is Associate Professor with the Department of Computer Science and Engineering, Southern University of Science and Technology (SUSTech), China. His research interests include machine learning, computer vision, and human-computer interaction.

\end{IEEEbiography}

\vfill

\end{document}